\documentclass[pdflatex,sn-nature]{sn-jnl}

\usepackage[T1]{fontenc}
\usepackage{graphicx}
\usepackage[sc]{mathpazo}
\renewcommand{\rmdefault}{ppl}
\usepackage{silence}
\usepackage{subcaption}
\usepackage{amsmath}
\usepackage{amssymb}
\usepackage{bbm}
\usepackage{booktabs}
\usepackage{multirow}
\usepackage{colortbl}
\usepackage{xcolor}
\usepackage{soul}
\usepackage{ulem}
\usepackage{enumitem}
\usepackage{wrapfig}
\usepackage{comment}
\usepackage{url}
\usepackage{pdfpages}
\usepackage{tabularx}
\usepackage{longtable}
\usepackage{needspace}
\usepackage{capt-of}
\usepackage{listings}
\usepackage[most]{tcolorbox}

\DeclareFontShape{T1}{cmtt}{bx}{n}{<->ssub * cmtt/m/n}{}
\AtBeginDocument{%
  \hypersetup{bookmarksdepth=3}%
  \pdfstringdefDisableCommands{%
    \def\markaug#1{#1}%
    \def\textit#1{#1}%
  }%
}

\newtcolorbox{promptbox}[1]{%
  colback=gray!3,
  colframe=black!40,
  title={\small\bfseries #1},
  fonttitle=\sffamily,
  breakable,
  enhanced,
  left=6pt, right=6pt, top=4pt, bottom=4pt,
  before upper={\footnotesize}%
}
\newtcolorbox{outputbox}[1]{%
  colback=blue!2,
  colframe=blue!30!black!40,
  title={\small\bfseries #1},
  fonttitle=\sffamily,
  breakable,
  enhanced,
  left=6pt, right=6pt, top=4pt, bottom=4pt,
  before upper={\footnotesize}%
}

\lstdefinestyle{appendixpython}{
    language=Python,
    basicstyle=\ttfamily\fontsize{5}{6}\selectfont,
    breaklines=true,
    breakatwhitespace=false,
    columns=fullflexible,
    keepspaces=true,
    showstringspaces=false,
    frame=single,
    framerule=0.2pt,
    xleftmargin=0pt,
    xrightmargin=0pt,
    aboveskip=0.4em,
    belowskip=0.2em
}

\providecommand{\burl}[1]{\begingroup\urlstyle{same}\nolinkurl{#1}\endgroup}

\sethlcolor{yellow}
\soulregister\cite7
\soulregister\ref7
\soulregister\textbf7
\soulregister\textit7
\soulregister\emph7
\soulregister\texttt7
\newcommand{\markaug}[1]{\textcolor{black}{#1}}

\title{\markaug{AI Scientists for Building Virtual-Cell Models}}

\author*[1]{\fnm{Xiangru} \sur{Tang}}\email{xiangru.tang@yale.edu}
\equalcont{These authors contributed equally to this work.}

\author[2,5]{\fnm{Jiapeng} \sur{Chen}}
\equalcont{These authors contributed equally to this work.}
\author[2,4,5]{\fnm{Yanjun} \sur{Shao}}
\equalcont{These authors contributed equally to this work.}

\author[1]{\fnm{Zhuoyun} \sur{Yu}}
\equalcont{These authors contributed equally to this work.}

\author[8]{\fnm{Yan} \sur{Cui}}
\author[6,7]{\fnm{Weixu} \sur{Wang}}
\author[10]{\fnm{Fang} \sur{Wu}}
\author[2]{\fnm{Yuxuan} \sur{Liao}}
\author[11]{\fnm{Yuchen} \sur{Zhuang}}
\author[12]{\fnm{Wenqi} \sur{Shi}}
\author[8,9]{\fnm{Zhi} \sur{Huang}}
\author[2]{\fnm{Arman} \sur{Cohan}}
\author[13]{\fnm{Xihong} \sur{Lin}}
\author[2]{\fnm{Smita} \sur{Krishnaswamy}}
\author[6,7]{\fnm{Fabian} \sur{Theis}}
\author*[1,2,3,4,5]{\fnm{Mark} \sur{Gerstein}}\email{mark.gerstein@yale.edu}

\affil[1]{\orgdiv{Department of Computer Science}, \orgname{Yale University}, \orgaddress{\city{New Haven}, \state{Connecticut}, \country{USA}}}

\affil[2]{\orgdiv{Program in Computational Biology and Biomedical Informatics}, \orgname{Yale University}, \orgaddress{\city{New Haven}, \state{Connecticut}, \country{USA}}}

\affil[3]{\orgdiv{Department of Molecular Biophysics and Biochemistry}, \orgname{Yale University}, \orgaddress{\city{New Haven}, \state{Connecticut}, \country{USA}}}

\affil[4]{\orgdiv{Department of Statistics and Data Science}, \orgname{Yale University}, \orgaddress{\city{New Haven}, \state{Connecticut}, \country{USA}}}

\affil[5]{\orgdiv{Department of Biomedical Informatics and Data Science}, \orgname{Yale School of Medicine, Yale University}, \orgaddress{\city{New Haven}, \state{Connecticut}, \country{USA}}}

\affil[6]{\orgdiv{Institute of Computational Biology}, \orgname{Helmholtz Munich}, \orgaddress{\city{Munich}, \country{Germany}}}

\affil[7]{\orgdiv{School of Computation, Information and Technology}, \orgname{Technical University of Munich}, \orgaddress{\city{Munich}, \country{Germany}}}

\affil[8]{\orgdiv{Department of Bioengineering}, \orgname{University of Pennsylvania}, \orgaddress{\city{Philadelphia}, \state{Pennsylvania}, \country{USA}}}

\affil[9]{\orgdiv{Department of Pathology and Laboratory Medicine, Perelman School of Medicine}, \orgname{University of Pennsylvania}, \orgaddress{\city{Philadelphia}, \state{Pennsylvania}, \country{USA}}}

\affil[10]{\orgdiv{Department of Computer Science}, \orgname{Stanford University}, \orgaddress{\city{Stanford}, \state{California}, \country{USA}}}

\affil[11]{\orgdiv{Department of Computer Science Engineering}, \orgname{Georgia Institute of Technology}, \orgaddress{\city{Atlanta}, \state{Georgia}, \country{USA}}}

\affil[12]{\orgdiv{Department of Genetics}, \orgname{Washington University in St. Louis}, \orgaddress{\city{St. Louis}, \state{Missouri}, \country{USA}}}

\affil[13]{\orgdiv{Department of Biostatistics}, \orgname{Harvard T.H.\ Chan School of Public Health}, \orgaddress{\city{Boston}, \state{Massachusetts}, \country{USA}}}

\abstract{Autonomous agents, from virtual laboratories to biological data-analysis agents and co-scientists, have begun to turn scientific goals into hypotheses. Whether they can go further and convert real, noisy and complex biological data into executable model hypotheses and architectures that encode testable assumptions remains unclear. Predicting how cells respond to genetic and chemical perturbations (i.e., modeling virtual cells) is a natural test of this ability: it is central to decoding disease mechanisms and therapeutic discovery, yet it demands cross-disciplinary expertise rarely concentrated in a single research group. Here, we introduce CellForge, an organized team of autonomous agents for single-cell perturbation modeling. Given a dataset and a task, it surveys the literature, parses the data, identifies relevant baselines, and formulates a model hypothesis through critic-mediated discussion among agents specialized in data modeling, single-cell biology, and deep learning. It then writes the training code, runs the experiments, and returns a trained predictor \markaug{including} architecture, code, and validation results. The architecture itself is the hypothesis: CellForge produces it end-to-end, with no human specifying the design, loss, or training protocol. Across six perturbation datasets spanning transcriptomic, epigenomic, and proteomic contexts, CellForge yields models that are competitive with hand-crafted state-of-the-art results on established tasks and extends the same workflow to paired multi-omic readouts.
\markaug{Blinded human evaluators} rate CellForge's research plans as more scientifically sound and innovative than those from single LLMs or deep-research systems.
\markaug{In summary, we show autonomous agents can complete the full hypothesis-to-validation cycle in single-cell biology, with an AI effectively building another AI, in a complex setting that currently requires a team of human experts.}
}

\keywords{virtual cell modeling, perturbation prediction, single-cell genomics, autonomous scientific discovery}

\begin{document}

\maketitle

\renewcommand{\figurename}{Fig.}

\section*{Introduction}
Modern biological research increasingly relies on high-dimensional perturbation datasets that span multiple intervention types and molecular readouts. Building a virtual cell that can predict cellular responses across these perturbations is an important objective for computational biology~\cite{bunne2024how} and has been formalized in the Virtual Cell Challenge~\cite{roohani2025virtual}. In parallel, the field is moving toward multimodal representation learning across transcriptomic, epigenomic, proteomic, and phenotypic measurements~\cite{cui2025multimodal}. Despite this progress, reliable generalization across unseen perturbation contexts remains limited, and the practical question of how to design transferable perturbation models is still unresolved.

The community has developed a rich toolkit for perturbation prediction, including conditional variational models~\cite{lotfollahi2019scgena,lotfollahi2023predictinga,piran2024disentanglement,lotfollahi2023predicting}, optimal transport~\cite{bunne2022supervised,bunne2023learning}, interactome-informed graph networks~\cite{roohani2024predictinga}, flow-based models~\cite{klein2025cellflow}, and transformers~\cite{cui2024scgpta,adduri2025predicting}, each demonstrating value on particular benchmarks. Yet no architectural choice transfers reliably across perturbation types, readout modalities, or evaluation protocols: performance claims depend strongly on split design and dataset composition~\cite{luecken2022defining,luecken2025defining,gao2024subtask,wei2025benchmarking,vinastorne2025systema,squair2024practical}, and deep models do not consistently outperform linear baselines under matched evaluation~\cite{ahlmanneltze2025deep}. Experimental scope has meanwhile expanded from transcriptomics to chromatin accessibility, paired RNA-protein, cytokine dynamics, and spatial context~\cite{liscovitchbrauer2021profiling,papalexi2021characterizing,gayoso2021jointa,ashuach2023multivia,saunders2025perturbmultimodal}, spanning genetic screens, chemical treatments, and cytokine responses across increasingly large atlases~\cite{adamson2016multiplexed,norman2019exploring,srivatsan2020massively,cui2024dictionary,peidli2024scperturba,zhu2025genomescale,zhang2025tahoe100m}, with cross-context and cross-modality modeling still maturing~\cite{yu2025perturbnet,li2025genetic,heumos2023best}. Each new setting has required rebuilding models from scratch, drawing on expertise rarely concentrated in a single group: the bottleneck is no longer data or compute, but the sustained effort of model design. A model hypothesis in this setting is an executable design: an architecture, loss, and training protocol that encode testable assumptions about cellular response. Here, we use \emph{a team of autonomous agents} to mean a coordinated set of large language model (LLM)-driven agents, each assigned a scientific or engineering role, that retrieves evidence, calls software tools, critiques proposals, and revises a shared research product with limited human intervention~\cite{gao2024empowering,swanson2024virtual,huang2025biomni}. Together, these trends define the challenge for an autonomous agent in perturbation biology: the system must read the literature, formulate a dataset-tailored model hypothesis, implement it, and evaluate it against well-accepted, community-defined splits. The relevant test is end-to-end completion in single-cell biology, rather than automation of isolated steps.

This challenge goes beyond any single architecture. It asks whether scientific reasoning about the model space itself can be automated: whether a system can formulate hypotheses grounded in biological literature, adapt them to each dataset, and instantiate them without manual design. Recent work shows that AI can assist parts of this cycle, from literature synthesis and hypothesis generation~\cite{si2025can,roohani2024biodiscoveryagent,schmidgall2025agentlaboratory,majumder2025discoverybench,panigrahi2026heurekabench} to autonomous software development~\cite{yang2024swe,li2022alphacode,wang2024openhands,jimenez2023swebench,chan2025mlebench}, with early applications in biomedicine spanning gene-set analysis, experiment automation, and single-cell exploration~\cite{wang2025geneagent,qu2026crisprgpt,huang2025biomni,swanson2024virtual,alber2026cellvoyager,penades2025ai}. Most recently, end-to-end autonomous AI research systems have demonstrated automated idea generation, code execution, and empirical evaluation in machine-learning settings~\cite{lu2026towards}. These systems clarify the operational template for an autonomous agent, but their benchmarks largely focus on self-contained computational tasks. Single-cell perturbation prediction is a sharper scientific testbed: each dataset requires data processing choices, single-cell biological assumptions, and deep-learning design to be reasoned about jointly, and an error in any one of them can invalidate an otherwise plausible architecture. Calls to extend autonomous AI research to such scientific domains~\cite{tang2025artificial,gao2024empowering,si2025can} have therefore outpaced concrete demonstrations. Existing systems still leave this gap on both sides: systems that reason over biology have not produced executable models~\cite{roohani2024biodiscoveryagent,alber2026cellvoyager}, whereas systems that generate and run code have not grounded their designs in biological deliberation~\cite{yang2024swe,li2022alphacode,wang2024openhands}. The full reasoning cycle from scientific hypothesis to trained, benchmarked model remains open, particularly for perturbation biology where model design choices interact tightly with evaluation protocol~\cite{luecken2025defining}.

Here we present \textsc{CellForge}, a team of autonomous agents for perturbation model development. To our knowledge, \textsc{CellForge} is the first system to close the design-to-implementation loop in this setting. Given a single-cell perturbation dataset and a task specification, its agents parse the data, benchmark existing baselines, survey the relevant literature, formulate a model hypothesis through structured multi-agent discussion, write and execute the training code, and return an executable model hypothesis: trained weights and an architecture specification, together with code and validation results. In this formulation, the model architecture itself is the hypothesis, proposed and validated end-to-end without human design intervention. We evaluate \textsc{CellForge} on six single-cell perturbation datasets (seven held-out prediction readouts) spanning genetic, chemical, and cytokine perturbations across scRNA-seq, scATAC-seq, and CITE-seq, asking whether autonomous hypothesis generation can substitute for manual model design across diverse perturbation settings. Crucially, the evaluation is fixed before model generation begins: tasks, held-out splits, and metrics follow community-defined criteria, so \textsc{CellForge}'s generated models are judged against standards external to the system rather than against protocols it selected or graded itself.

\section*{Results}

\subsection*{CellForge turns a biological task into an executable model hypothesis}

Predicting how cells respond to a perturbation has become a central computational problem in single-cell biology, yet every new perturbation atlas currently demands a bespoke predictor, designed by a team that spans cell biology, statistics, and deep learning. The first test for CellForge is therefore whether it can replace this manual model-design step by producing, from a concrete perturbation-prediction task, not only a written proposal but an executable model hypothesis (Figure~\ref{fig:intro_perturbation}a,b). Given a single-cell perturbation dataset and a task description, CellForge returns a trained model together with its architecture specification and training code. In this setting, the model hypothesis is the executable predictor itself: its architecture, loss, and training protocol encode assumptions about how perturbations change cell state. We test these assumptions across the perturbation classes and molecular readouts summarized in Figure~\ref{fig:intro_perturbation}c, treating candidate model architectures as hypotheses to be selected and evaluated empirically (Figure~\ref{fig:intro_perturbation}d).

CellForge turns the task into this executable hypothesis through three stages (Figure~\ref{fig:overview}a). \markaug{Formulation fixes the empirical problem before any new architecture is proposed; then during Ideation,} expert proposals undergo critic-mediated review and revision; the selected research plan then passes to coding agents for iterative implementation and execution (Figure~\ref{fig:overview}b). \markaug{In more detail,} (1) \markaug{\textbf{Formulation}} parses the perturbation, feature, and context fields; constructs held-out perturbation splits; reproduces established baselines from previous papers; and locks the metric pipeline into a task report. (2) \textbf{Ideation} uses this task report as the basis for model design. Data modeling, single-cell biology, and deep learning agents retrieve literature, propose candidate designs, critique one another's assumptions, and revise the preprocessing strategy, neural architecture, loss function, and training protocol through critic-mediated discussion, with coordination trajectories tracked across rounds (Figure~\ref{fig:overview}c). The protocol mirrors conference peer review: domain experts first-author separate proposals and then review one another's plans. A central critic acts like an area chair, synthesizing the reviews into a meta-review that guides revisions and final selection. (3) \textbf{Execution} treats the approved plan as an implementation target: it writes runnable training code, launches five-fold cross-validation, repairs failed implementations, and returns final model weights, an architecture specification, source code, and validation metrics.

We applied the same formulation--ideation--execution loop to six perturbation-response datasets spanning scRNA-seq, paired RNA-protein CITE-seq, and scATAC-seq. These experiments test whether CellForge can design models that predict cell responses to unseen perturbations, a setting where generalization has traditionally been difficult. The remainder of the Results therefore proceeds in three steps: first, we summarize the architectural motifs selected by CellForge; next, we test performance on established scRNA-seq benchmarks with published reference methods; and finally, we evaluate under-studied perturbation settings where task-specific predictors are limited.

Before comparing predictive performance, we first inspected whether generated architectures varied with the biological task. Across tasks, the architectural motifs selected by CellForge exhibited clear task-specific patterns rather than converging on a single template. Among the established scRNA-seq benchmarks, single-gene CRISPRi models primarily used MLP and attention components, combinatorial CRISPRa models more often included attention and bottleneck components, and drug-response models showed the widest range of motifs, including VAE-style latent encoders, optimal-transport modules, and normalizing-flow components. The under-studied settings extended this pattern: temporal cytokine models used VAE and MLP-mixing components suited to continuous response trajectories; CITE-seq models used graph-attention and Transformer components to combine paired RNA and protein readouts; and scATAC-seq models used diffusion-based denoising modules suited to sparse, near-binary chromatin accessibility data. The next two sections examine these dataset-specific architectures in detail and evaluate their empirical performance.

\subsection*{CellForge designs competitive architectures on established scRNA-seq benchmarks}

CellForge did not converge on a single universal architecture. Instead, the generated models showed task-specific patterns that often aligned with the structure of each perturbation problem. We first considered three established scRNA-seq benchmarks for which published reference methods are available: single-gene CRISPRi~\cite{adamson2016multiplexed}, combinatorial CRISPRa~\cite{norman2019exploring}, and drug perturbation~\cite{srivatsan2020massively}. For genetic perturbations, the published references included CPA, scGen, CondOT, Biolord and scGPT~\cite{lotfollahi2023predictinga,lotfollahi2019scgen,bunne2022supervised,piran2024disentanglement,cui2024scgpt}. For drug perturbations, they included ChemCPA and CellFlow~\cite{hetzel2022predicting,klein2025cellflow}, alongside unperturbed, random-forest and linear-regression controls. We reproduced the applicable methods and evaluated them with CellForge using the same five held-out-perturbation folds and metric pipeline (Methods). Figure~\ref{fig:scrna_models}a--c illustrates the three CellForge architectures, with full implementation details provided in Supplementary Section~\ref{app:designed_models}. \markaug{Figure~\ref{fig:scrna_models}d summarizes global and differential-feature prediction, while panels e and f assess differential-feature recovery through receiver operating characteristic (ROC) curves and the corresponding area under the ROC curve (AUROC), and panel g aggregates ranks across the same metric families.} Differential-feature recovery tests whether features ranked by absolute predicted change recover the top 50 features in the observed perturbation-versus-control comparison (Methods).

For single-gene CRISPRi, CellForge proposed \textit{CPA-X}, an MLP expression encoder coupled to a multi-head cross-attention block in which a perturbation embedding attends to the encoded control-cell state. This design is related to ideas in scGen and CPA~\cite{lotfollahi2019scgen,lotfollahi2023predictinga}, and is well matched to a setting where each example is defined by one genetic perturbation and one observed expression response. Empirically, CPA-X achieved the highest Pearson correlation on the Adamson benchmark, indicating strong recovery of global expression structure. Its differential-expression error was higher than that of the best reference methods, suggesting that global expression agreement and differential-gene fidelity remain partly separable objectives.

For combinatorial CRISPRa, CellForge generated \textit{scGen-X}, a bottlenecked expression encoder with a cross-attention module that fuses perturbation identity with cell context before a residual MLP decoder. This architecture resembles design principles used in GEARS and AttentionPert~\cite{roohani2024predicting,bai2024attentionpert}, and is consistent with the need to model interactions between paired perturbations. On the Norman benchmark, scGen-X achieved the lowest differential-expression mean squared error, showing that the generated design was competitive in a setting where combinatorial effects are central to the task.

For drug response, CellForge generated \textit{ChemCellFlow}, which combines gene and chemical encoders, conditions on compound identity and dose, applies a Sinkhorn-based conditional optimal-transport block, and refines the latent state with a normalizing-flow module. This architecture is closest to CellFlow and related neural optimal-transport approaches~\cite{klein2025cellflow,bunne2023learning,hetzel2022predicting}, suggesting that CellForge recovered the main design principle of a strong human-designed model: treating drug response as a conditional transport problem between cellular states. On the Srivatsan drug benchmark, CellFlow~\cite{klein2025cellflow}, which was developed and tuned specifically for this setting, achieved the best performance. ChemCellFlow achieved the second-best result, outperforming the remaining reference methods despite using a simpler coupling of chemical, dose, and cellular state representations. Thus, even when the best method was a specialized human-designed model, CellForge identified the relevant architecture family and automatically produced a competitive implementation.

Together, these results show that CellForge can recover strong architecture families for established scRNA-seq perturbation benchmarks, with performance competitive with published reference methods while preserving task-specific model structure.

\subsection*{CellForge extends model design to under-studied perturbation settings}

We next asked whether CellForge could design models beyond the established expression-only endpoint benchmarks above. We considered three settings where task-specific reference predictors are limited: cytokine-driven differentiation dynamics in the Schiebinger dataset~\cite{schiebinger2019optimal}, paired RNA-protein perturbation readouts in the Papalexi CITE-seq dataset~\cite{papalexi2021characterizing}, and chromatin-accessibility perturbation in the Liscovitch-Brauer scATAC-seq dataset~\cite{liscovitchbrauer2021profiling}. These tasks test whether an autonomous design system can adapt to time-dependent responses, multimodal readouts, and sparse regulatory measurements. To our knowledge, no manually crafted, published task-specific prediction models are available for these datasets. We therefore benchmarked CellForge against a common reference set of four established model types (linear regression, random forest, \markaug{graph neural network and Transformer}), all trained and evaluated under an identical protocol. Figure~\ref{fig:understudied_models}a--c illustrates the representative CITE-seq, scATAC-seq, and cytokine architectures, respectively, with full implementation details provided in Supplementary Section~\ref{app:designed_models}.

For cytokine-driven differentiation, CellForge generated \textit{CPA-Traj}, a trajectory-aware VAE with shared and condition-specific latents, a continuous time covariate, and a perturbation-mixing module before an MLP decoder. This design reflects a setting where perturbation effects unfold along a continuous response trajectory rather than as a single endpoint. CPA-Traj ranked first on several global and differential-feature metrics (Figure~\ref{fig:understudied_models}d). Differential-feature recovery is reported using ROC curves and AUROC in Figure~\ref{fig:understudied_models}e,f, respectively.

For paired RNA-protein CITE-seq, CellForge used the joint architecture \textit{totalGAT} shown in Figure~\ref{fig:understudied_models}a. The title-case label \textit{TotalGAT} in panel a denotes the same model. We refer to it as \textit{totalGAT} throughout the text. \textit{totalGAT} used modality-specific encoders, graph-attention layers, cross-attention with perturbation embeddings, and separate output heads for RNA and protein. This design reflects the need to combine two molecular readouts while preserving distinct output spaces. On the Papalexi dataset, CellForge achieved the lowest MSE for both RNA and protein prediction and improved RNA differential-expression correlation relative to the generic baselines, although generic neural baselines remained competitive on some global correlation metrics. \markaug{Figure~\ref{fig:understudied_models}d shows global and differential-feature prediction for both CITE-seq readouts, with differential-feature recovery summarized by ROC curves and AUROC in panels e and f and joint rankings in panel g.} Recovery was weaker for protein than for RNA, consistent with the difficulty of predicting perturbation-sensitive protein features. This pattern indicates partial recovery of paired-modality structure rather than uniformly strong performance across both molecular readouts.

For scATAC-seq, CellForge generated \textit{ChromDDPM}, a DDPM-style~\cite{ho2020ddpm} denoising model conditioned on cell state and perturbation information. This design is a natural fit for sparse, near-binary chromatin-accessibility profiles, where prediction can be viewed as recovering perturbation-sensitive accessibility signal from a noisy high-dimensional readout. On the Liscovitch-Brauer dataset, overall performance was lower than in the expression-based tasks, but CellForge improved over simple baselines on several metrics (Figure~\ref{fig:understudied_models}d,e,g). At the same time, the modest global performance and recovery curves indicate partial rather than complete recovery of regulatory response structure.

Together, these results show that CellForge can extend beyond well-studied scRNA-seq benchmarks. In settings where task-specific perturbation predictors are scarce, it assembled established machine-learning components into modality-aware architectures that were often competitive with generic neural baselines and generally stronger than simple learners, without manual architecture engineering.

\subsection*{Blinded evaluators prefer CellForge's research plans over single-agent and deep-research alternatives}

The benchmark results above evaluate final trained models, but they do not by themselves show whether CellForge improves the quality of the research plans that precede code execution. To test this directly, we compared CellForge plans with plans generated from identical task specifications by OpenAI Deep Research~\cite{openai2025introducing}, Perplexity Deep Research~\cite{perplexity2025deepresearch}, Gemini Deep Research~\cite{google2025geminideepresearch}, Biomni~\cite{huang2025biomni}, and a single-LLM prompting baseline. All plans were evaluated blind to system identity. An example trace shows the conference-review-like process used to generate a CellForge plan: experts propose and review plans, and an AC-like critic provides a meta-review that guides revision (Figure~\ref{fig:llm-eval}a).

In the LLM-judge evaluation, CellForge achieved the highest mean score in 14 of 15 task--criterion combinations (Figure~\ref{fig:llm-eval}b); OpenAI Deep Research scored higher only for technical feasibility on the drug task. A separate blinded evaluation by three single-cell biology experts assessed technical novelty, feasibility, clarity and consistency, biological plausibility, mechanism explanation, pathway relevance, and cross-perturbation generalizability (Figure~\ref{fig:llm-eval}c). This human evaluation likewise favored CellForge across most task--criterion combinations, although not uniformly. Pairwise correlations among the five LLM judges ranged from 0.88 to 0.93, \markaug{indicating strong agreement among their plan evaluations} (Figure~\ref{fig:llm-eval}d). Mean human-expert scores correlated more strongly with the corresponding LLM-judge scores than with CellForge's internal graph-based discussion confidence score, \markaug{indicating that the external LLM judges more closely tracked human assessments than did the system's internal deliberation signal} (Figure~\ref{fig:llm-eval}e). Phase-resolved profiles further show how the systems were judged on task analysis and method design across the three perturbation settings (Figure~\ref{fig:llm-eval}f). Together, these complementary LLM and human reviews provide plan-level evidence that the multi-agent deliberation improved judged plan quality before code execution. The human panel was small, however, and neither evaluation constitutes independent biological validation or evidence about final trained-model performance.

\section*{Discussion}

CellForge demonstrates that model design itself can be treated as a scientific hypothesis: rather than selecting from predefined architectures, an autonomous team reasons over the literature, formulates dataset-tailored modeling hypotheses, and instantiates them as trained, benchmarked models. The resulting architectures were synthesized through scientific reasoning conditioned on each dataset's characteristics rather than drawn from a fixed search space, and produced competitive predictions across genetic, chemical, and cytokine perturbation benchmarks, including the highest Pearson correlation on Adamson CRISPRi ($PCC = 0.9883$) and the lowest differential-expression error on Norman CRISPRa ($MSE_{\text{DE}} = 0.1736$). Prior agentic systems for science occupy distinct, narrower roles. Virtual Lab~\cite{swanson2024virtual} orchestrates existing tools such as ESM and Rosetta into pipelines but does not invent architectures. CellVoyager~\cite{alber2026cellvoyager} generates hypotheses from biological data but stops short of implementation. \textit{The AI Scientist}~\cite{lu2026towards} automates template-based ML ideation and coding but evaluates by self-review and is not grounded in domain literature. CellForge closes this gap by treating model architecture itself as a hypothesis to be reasoned over, implemented, and evaluated against fixed, community-defined benchmark splits that precede model design, an inversion that addresses the reproducibility problem in computational biology where evaluation protocol and model development are otherwise entangled. A further distinction is that \textit{The AI Scientist} validates outputs through LLM-generated self-review, whereas CellForge validates against held-out perturbation conditions with experimentally measured single-cell readouts, a requirement that is not negotiable in the single-cell setting. Together these properties establish that autonomous research can produce computational methods validated on the same held-out perturbation splits as the human-designed methods they compete against.

While CellForge produces competitive perturbation models, the current system has several limitations. Three runs per dataset is too few to characterize the full distribution of architectures CellForge would generate across many runs. More broadly, recent evidence suggests that AI tools expand individual scientific impact while contracting science's collective focus~\cite{hao2026artificial}; autonomous systems like CellForge may be subject to the same dynamic, as LLM-driven ideation can systematically favor architectures that are well-represented in training data regardless of dataset-specific conditioning, and whether the critic-mediated deliberation genuinely diversifies architectural choices across datasets has not been directly measured. Human review is intentionally minimal, limited to two steps: a brief review of research plan plausibility before execution begins, and training-script compliance before any GPU job is launched, to prevent harmful or resource-wasteful execution. Whether this two-step design represents the right minimal footprint and whether it generalizes to more complex task specifications have not been tested directly. Validation here is also computational: CellForge's predictions are scored against held-out perturbations measured experimentally by the original authors, but the framework has not yet been coupled to a wet-lab loop, and prospective experimental validation of its predictions remains future work. Spatially resolved perturbation transcriptomics is a natural next testbed for the framework itself. Such datasets replace unordered cell sets with spatial graphs and may require different model classes~\cite{saunders2025perturbmultimodal,binan2025simultaneous,shen2026spatialperturbseq}, and no dominant modeling paradigm has yet emerged for them.

The broader significance of CellForge is that it attempts to design models, not just assemble pipelines or analyze data. In this setting, model architecture is itself a scientific hypothesis. Choices about the encoder reflect assumptions about how perturbation signals move through gene regulatory space, whereas choices about the decoder reflect assumptions about the structure of gene expression manifolds. The next question is whether the agents' explanations for these choices match the mechanisms that actually drive model performance. Figure~\ref{fig:llm-eval} shows that blinded LLM judges and a separate panel of single-cell biology experts generally preferred CellForge's research plans. The strong correspondence between human and LLM scores supports the use of LLM judges for plan-level comparison, whereas their weaker correspondence with CellForge's internal confidence score indicates that deliberation confidence is not itself a substitute for external evaluation. The agents reason meaningfully about the task, but stronger tests will be needed to determine whether that reasoning is mechanistically faithful. Perturbation prediction is a useful proving ground because it is both well benchmarked and a rapidly advancing area of single-cell machine learning, making proposed designs straightforward to compare against strong baselines. More broadly, autonomous method design may prove most valuable in ill-defined biological settings, where no dominant modeling paradigm yet exists and where a human team would otherwise spend months on architecture exploration before reaching a working baseline.

\section*{Methods}

\textsc{CellForge} takes as input a processed single-cell perturbation dataset and a task description; the pipeline is grounded by a community-agreed evaluation criterion fixed before any architecture is proposed and a seed corpus of key papers used to initialize literature retrieval. It returns a trained model with its design (model weights, architecture specification, and training code). A team of autonomous agents parses the data, benchmarks existing baselines, retrieves additional literature, formulates a model hypothesis through multi-agent discussion, generates code, trains the model, and reports results, across three sequential stages: \emph{Formulation}, \emph{Ideation}, and \emph{Execution} (Figure~\ref{fig:overview}). Human review is restricted to two steps: research plan plausibility before execution begins, and training-script compliance before any GPU job is launched. Full prompt templates and run logs are released with the codebase.

\subsection*{Overview}

\vspace{-.2cm}
\paragraph{Problem formulation.}
We study held-out perturbation prediction. For a control-state molecular profile $x$, perturbation identity $p$, and optional context $c$ (for example cell type, batch, or donor), the goal is to learn a predictor $f_{\theta}(x,p,c) \mapsto y$ that returns the post-perturbation molecular profile $y$ for perturbations not observed during training. Prediction is typically mediated by a learned encoder $g_{\phi}(x)=z$ that maps cells into a latent space before applying perturbation-specific transformations; however, the encoder, decoder, and loss are designed per dataset rather than fixed in advance.
Perturbation identities are split into five folds, and we evaluate all methods only on held-out perturbations. We score all baselines and CellForge-designed models under the same split definition and metric pipeline. For deterministic baselines, the reported score is the average over folds. Each CellForge run comprises a full design-implement-train cycle evaluated across the same five folds. Summary values reported in the manuscript are averaged over three independent runs.

\paragraph{Evaluation metrics.}
Following recent perturbation benchmarks and evaluation tooling~\cite{roohani2024predicting,adduri2025predicting,celleval2026}, metric computation is perturbation-centric: scores are computed per held-out perturbation rather than per cell. For each held-out perturbation (and matched context, when context labels are available), we first aggregate predicted cells and observed cells separately into mean profiles over the shared feature space. We then compute mean squared error (MSE), Pearson correlation (PCC), and the coefficient of determination ($R^2 = 1 - \text{SS}_{\text{res}}/\text{SS}_{\text{tot}}$, which can be negative when predictions are worse than the mean) between the predicted and observed profiles across features for that perturbation. For visualization, MSE is transformed as $-\log_{10}(\max(\mathrm{MSE},10^{-12}))$, so larger values consistently indicate lower error. Negative values are expected when MSE exceeds 1 and do not denote a negative error. Because MSE depends on the scale of each molecular readout, its absolute transformed value should be compared among methods within a dataset, not across datasets. Reported summary values are obtained by averaging these per-perturbation scores over all held-out perturbations and then across folds. Because these global metrics are dominated by highly abundant and hub features that vary little across perturbations, they can overstate predictive accuracy for perturbation-specific effects. We therefore also compute $\mathrm{PCC}_{\mathrm{DE}}$, $\mathrm{MSE}_{\mathrm{DE}}$, and $R^2_{\mathrm{DE}}$ over the observed differentially expressed genes for each perturbation.

\paragraph{DEG recovery metrics.}
For each held-out perturbation and molecular readout, we rank observed genes using \texttt{scanpy.tl.rank\_genes\_groups}. Genes are ordered by the absolute test statistic, with \texttt{rankby\_abs=True} and \texttt{use\_raw=False}. The established scRNA-seq benchmarks use Scanpy's \texttt{t-test\_overestim\_var} option. The under-studied multimodal benchmarks use the Wilcoxon option. The top 50 observed differentially expressed genes define the positive set. For a readout with $p$ shared genes, its size is $\min(50,\lfloor p/2\rfloor,p-1)$. These sets are fixed within each evaluation fold and shared across all methods and CellForge runs. They are used only for post hoc evaluation.

For every method, predicted genes are ranked by the absolute difference between the predicted perturbation mean and the observed control mean. Recovery@50 is labelled ``Recall'' in the ranking heatmaps. It is the overlap between the top 50 observed and predicted differentially expressed genes, divided by the number of genes compared. ROC curves use membership in the observed differentially expressed gene set as the binary outcome. The absolute predicted change is the continuous score across all shared genes, and AUROC summarizes the curves.

\paragraph{Human involvement.}
Human review is intentionally minimal and limited to two steps. The first occurs after Ideation: a reviewer confirms that the research plan is scientifically plausible and internally coherent, covering dataset parsing, split definition, and benchmark feasibility, without modifying the plan content. The second occurs before any GPU job is launched: the reviewer checks that the training script matches the approved plan and complies with local HPC usage policy, to prevent harmful or resource-wasteful execution. Failed checks at either step trigger regeneration rather than manual editing.

\paragraph{Blinded planning evaluation.}
To separate plan quality from downstream code execution, we evaluated anonymized CellForge and baseline research plans using two blinded review procedures. A research plan is the Ideation-stage output and specifies task analysis, preprocessing, neural architecture, loss function, and training protocol. First, five independent LLM judges (Claude 3.7, DeepSeek-R1, OpenAI o1, Qwen-2.5, and Llama 3.1) scored each plan from 1 to 10 for scientific validity, technical feasibility, innovation, experimental design, and impact potential. We removed formatting cues to reduce style bias and randomized plan order in each evaluation round. Scores were averaged across five independent rounds, and pairwise Pearson correlations were used to quantify inter-judge reliability. Second, three domain experts in single-cell biology performed a separate blinded review of the same anonymized plans, scoring technical novelty, feasibility, clarity and consistency, biological plausibility, mechanism explanation, pathway relevance, and cross-perturbation generalizability. We used Pearson correlation to compare mean human-expert scores with the corresponding aggregate LLM-judge scores and, for CellForge proposals, with the graph-based discussion confidence score. For phase-resolved visualization, task-analysis scores summarized dataset analysis, task-type analysis, and baseline-defect analysis, whereas method-design scores summarized scientific validity, technical feasibility, innovation, experimental design, and impact potential. The human review did not alter the plans or participate in CellForge's design loop.

\subsection*{Formulation}

The Formulation stage converts the processed dataset, task description, and community-agreed evaluation criterion into a well-defined research question and standardized benchmark setting that grounds all subsequent design and evaluation decisions.

\paragraph{Data parsing and benchmark definition.}
The system reads processed expression matrices (cell-by-feature count tables) across modalities (scRNA-seq, scATAC-seq, CITE-seq). The upstream step of converting raw sequencing reads into count matrices has already been automated by agentic pipelines~\cite{youngblut2025scbasecamp} and could be composed with CellForge to span the full path from FASTQ to trained model; we do not pursue this here, as all benchmarks used in this study provide expression-level data. From these matrices, the system identifies perturbation, feature, and context columns, extracts the summary statistics needed to define a valid prediction task, and constructs a unified five-fold split over perturbation identities. It also checks that each fold preserves sufficient cellular coverage for training and testing. This component outputs a standardized dataset report and split definition; a representative parsed summary and the associated prompt structure are provided in Supplementary Sections~\ref{app:data_parser}.

\paragraph{Baseline benchmarking.}
The Formulation stage consults the seed papers supplied as an input to identify baseline families relevant to the task specification, reproduces those baselines under the same five-fold split, and records a reference benchmark before any new architecture is designed. Because baseline comparisons are sensitive to implementation details, a human reviewer verifies that the selected baselines, code paths, and evaluation setup are appropriate for the task before these reference results are used downstream. This stage outputs the task report, split definition, and baseline performance context used by the Ideation stage.

\subsection*{Ideation}

The Ideation stage translates the task report, retrieval results, and baseline evidence into a concrete, literature-grounded modeling plan through structured expert discussion. The goal is to identify a dataset-matched design space, not to tune hyperparameters of a fixed template.

\paragraph{Literature retrieval.}
At the start of research ideation, the retrieval system combines the seed corpus of key papers provided as an input (46 papers across the benchmarks reported here) with dynamic PubMed and web searches that extend the seed set. The local corpus is indexed in Qdrant for embedding-based nearest-neighbor lookup and provides stable coverage of benchmark papers, perturbation-response models, multimodal methods, foundation models, and implementation resources. Candidate documents are ranked by cosine similarity over Sentence-BERT embeddings~\cite{reimers2019sentence}, and the search alternates between breadth-first exploration and depth-first follow-up so that the retrieved set covers both diverse modeling families and practical implementation details. We use this component as a retrieval heuristic rather than evaluate it as a standalone retrieval benchmark. In practice, the retrieval stage applies duplicate pruning and minimum relevance thresholds to control search breadth while retaining implementation-specific evidence. The full fixed-corpus inventory is provided in Supplementary Section~\ref{app:knowledge_base}.

\paragraph{Critic-mediated expert panel.}
Within any single run, all experts share the same LLM backend and differ only in their role-specific system prompts, so that design variation reflects the multi-agent protocol rather than differences between foundation models. Across experiments, we evaluated five LLM backends under matched decoding parameters. For a given task, the system selects a subset of experts such as a data-modeling expert, a single-cell biology expert, and a deep-learning expert, together with a critic that reads all proposals and provides structured feedback. Operationally, experts do not write code in this stage; they iteratively refine model plans, including encoder structure, perturbation representation, loss design, and training strategy. For each expert's current proposal, the panel tracks a coordination score $c_t^{(i)} \in [0,1]$, where i indexes the experts and t the discussion round; it guides deliberation and stopping but is neither a calibrated probability nor an experimental performance measure. The protocol mirrors conference review. Each expert first acts as an author by submitting a plan, then as a reviewer by assessing the other plans. The central critic acts as an area chair, synthesizing peer reviews into a meta-review and score for each plan. Experts revise their proposals in the next round, and only the selected final plan advances to Execution.

\paragraph{Discussion protocol.}
The discussion implements this conference-review process over at most $T_{\max}=20$ synchronous rounds. Each round has four steps. First, every expert acts as an author and submits a revised plan $m_t^{(i)}$, conditioned on the task report, retrieved evidence, and earlier feedback. Second, every other expert $j$ (where $j \neq i$) acts as a reviewer, returning written comments and a peer score $r_{\mathrm{peer},t}^{(i,j)} \in [0.1,1.0]$. Third, the central critic acts as an editor or area chair. It reads the proposals and peer reviews, evaluates scientific merit, technical feasibility, biological validity, comparison with published methods, and implementation strategy, and issues a meta-review with an overall score $r_{\mathrm{crit},t}^{(i)} \in [0.1,1.0]$. Fourth, the authors receive the peer reviews and meta-review before revising their plans for the next round. These scores are LLM judgments produced by role-specific review prompts, not measurements on held-out data. The panel combines them in a fixed weighted moving average,
\begin{equation}\label{eq:confidence}
c_t^{(i)} = 0.3\, c_{t-1}^{(i)} + 0.4\, r_{\mathrm{crit},t}^{(i)} + \frac{0.3}{k-1}\sum_{j \neq i} r_{\mathrm{peer},t}^{(i,j)}.
\end{equation}
Here $c_0^{(i)}$ is the initial score before review, which is 0.5, and $k$ is the number of experts. The update assigns 40\% weight to the critic's editorial assessment, 30\% to the mean peer assessment, and 30\% to the previous-round score. It ranks proposals by panel deliberation rather than held-out predictive performance. After each round, the coordination scores identify weaker proposals and guide the critic's next meta-review and the authors' revisions. All candidate plans remain under discussion until every proposal has $c_t^{(i)} \geq 0.8$ and the largest pairwise score difference is below $0.03$. If this convergence criterion is not met by $T_{\max}$, the process stops at the round limit. At termination, the critic uses the final ranking and accumulated reviews to make an editor-like decision. Lower-ranked alternatives are excluded, and only the selected final plan is incorporated into the modeling brief and advanced to Execution.

The Ideation stage outputs the task report, split definition, baseline performance context, and an approved research plan (preprocessing strategy, neural architecture, loss function, and training protocol) that together define a dataset-matched design space for the Execution stage.

\subsection*{Execution}

The Execution stage converts the approved research plan into runnable code, submits approved training jobs, and validates results. A detailed compute profile and cost estimation, including token counts, GPU hours, and a provider-agnostic price-conversion formula, is provided in Supplementary Section~\ref{app:cost_analysis}.

\paragraph{Code generation and repair.}
We implement code generation with OpenHands~\cite{wang2024openhands}, a software-engineering agent that iteratively proposes actions, executes them in a sandboxed workspace, observes the resulting outputs, and revises subsequent actions. Given the approved research plan, the Code Generator writes Python training code with pinned dependencies, executes the code, inspects runtime feedback, applies patches, and re-executes. Code generation follows a ReAct-style loop: propose action, execute, observe output, revise~\cite{yao2023react}. This loop repeats for up to 5 attempts. If all attempts fail, the system rolls back to the last passing state; runs that fail to produce a trainable model after exhausting all repair attempts are excluded from the reported results and replaced by a new independent run, so that all summary statistics reflect three successfully completed design-train cycles.
A breakdown of resolved error categories is provided in Supplementary Section~\ref{app:failure_case_analysis}.

\paragraph{Training and validation loop.}
After approval, the scheduler launches training with 5-fold cross-validation, early stopping (patience = 10 epochs), a cosine-annealing learning-rate schedule, and periodic checkpointing. Within each fold, early stopping monitors validation loss every epoch and halts training if no improvement is observed for 10 consecutive epochs. After each fold, the Validation Agent evaluates the best checkpoint with the perturbation-level metric pipeline defined above and summarizes failure modes such as systematic under-prediction of large-magnitude responses or instability on specific perturbation families. Across folds, the Validation Agent applies a second-level check: if validation loss increases for 3 consecutive folds, it adjusts regularization strength or reduces the number of training epochs and restarts from the best checkpoint. These validation summaries are then fed back to the Code Generator for a subsequent repair round.

\subsection*{Data availability}
All datasets used in this study are publicly available. The Adamson et al.~\cite{adamson2016multiplexed}, Norman et al.~\cite{norman2019exploring}, and Srivatsan et al.~\cite{srivatsan2020massively} datasets are available at NCBI GEO under accessions GSE90546, GSE133344, and GSE139944, respectively. Pre-processed \texttt{.h5ad} files for these datasets are also available through the scPerturb resource~\cite{peidli2024scperturba} under DOI \texttt{10.5281/zenodo.13350497}.

\subsection*{Code availability}
The source code of CellForge is publicly available at \texttt{github.com/gersteinlab/CellForge}. All reported model training runs were executed through Slurm on Yale Center for Research Computing infrastructure, primarily on the \texttt{misha} and \texttt{bouchet} clusters, with production jobs submitted via \texttt{sbatch}.

\backmatter

\subsection*{Funding}
This work was supported by the ALW Professorship Funds.

\subsection*{Author contributions}
X.T., J.C., Y.S., and Z.Y. contributed equally to this work. X.T., J.C., Z.Y., Y.S., and Y.C. conceived the study. X.T. led the project. J.C., Z.Y., and Y.S. developed the CellForge multi-agent framework. J.C. implemented and benchmarked all single-cell baseline methods. J.C., Z.Y., Y.S., and Y.C. analyzed and interpreted model outputs. Y.S., J.C., and Z.Y. created the figures. W.W. contributed to experimental design and downstream analysis directions. F.W., Y.Zh., W.S., Z.H., A.C., X.L., S.K., and F.T. reviewed the manuscript and provided feedback. X.T. and M.G. supervised the project. Y.S., J.C., X.T., and Z.Y. wrote the manuscript. All authors read and approved the final manuscript.

\subsection*{Acknowledgements}
The authors thank the Yale Center for Research Computing for providing computational resources and technical support.

\subsection*{Competing interests}
The authors declare no competing interests.

\subsection*{Additional information}
Supplementary Information is available for this paper. Correspondence and requests for materials should be addressed to X.T. or M.G.

\normalem
\bibliography{reference}

\clearpage
\clearpage
\section*{Figure legends}
\setcounter{figure}{0}

\refstepcounter{figure}
\label{fig:intro_perturbation}
\noindent\textbf{Fig. \thefigure{} \textbar{} CellForge as an autonomous virtual-cell model designer.}
\textbf{a}, Perturbation-prediction task. Given a control cell state $X_{\mathrm{ctrl}}$ and a perturbation drawn from perturbation space $\mathcal{P}$, the model predicts the corresponding post-perturbation state $X_{\mathrm{pert}}$.
\textbf{b}, CellForge inputs and outputs. User-specified inputs include a preprocessed single-cell perturbation dataset, a perturbation task description, an evaluation criterion, and selected seed papers. CellForge operates as a team of autonomous agents and returns model weights, an architecture specification, and source code. The generated architecture is treated as the design hypothesis because CellForge produces the architecture, loss, and training procedure end-to-end rather than receiving them from human designers.
\textbf{c}, Perturbation types and molecular readouts considered in this study. The benchmark covers drug, gene-editing, and cytokine perturbations, with readouts from scRNA-seq, scATAC-seq, and paired RNA--protein CITE-seq.
\textbf{d}, Conceptual view of model architecture as a hypothesis. Existing architectures and newly generated modeling hypotheses occupy a task-dependent design space, within which CellForge identifies a candidate architecture for implementation and empirical evaluation.
\par\vspace{1.25em}

\refstepcounter{figure}
\label{fig:overview}
\noindent\textbf{Fig. \thefigure{} \textbar{} CellForge workflow for formulation, ideation, and execution.}
\textbf{a}, Task-to-hypothesis workflow. Given a biological modeling task, CellForge develops a corresponding modeling hypothesis through three stages. \emph{Formulation} performs dataset analysis, task overview, baseline reproduction, and benchmark definition to produce a task-analysis report. \emph{Ideation} retrieves literature and coordinates iterative expert discussion to produce a research plan. \emph{Execution} generates code, trains models, and iteratively refines implementations to produce the final model, code, and results.
\textbf{b}, Research-plan discussion and code implementation. Data-modeling, single-cell biology, and deep-learning experts propose candidate plans. A critic evaluates the proposals, and confidence-guided peer discussion and revision produce a final plan specifying preprocessing, model architecture, and loss. Coding agents then implement and execute the plan, iteratively refine candidate implementations, and return executable code, model checkpoints, and predicted responses.
\textbf{c}, Confidence-score trajectories during ideation for the Norman, Adamson, and Srivatsan benchmarks. Scores are shown by discussion round for the data-modeling expert, single-cell biology expert, deep-learning expert, model-training expert, and critic agent, with divergence, convergence, and stable phases indicated. Each score aggregates the previous-round score, critic assessment, and peer assessment; it is neither a calibrated probability nor a predictive-performance measure.
\par\vspace{1.25em}

\refstepcounter{figure}
\label{fig:scrna_models}
\noindent\textbf{Fig. \thefigure{} \textbar{} CellForge architectures benchmarked on established scRNA-seq perturbation-prediction tasks.}
\textbf{a}, \textit{CPA-X} architecture for single-gene CRISPRi, in which an MLP expression encoder generates context embeddings and a perturbation embedding queries the encoded control-cell state by cross-attention before residual MLP decoding.
\textbf{b}, \textit{scGen-X} architecture for combinatorial CRISPRa, in which a bottlenecked expression encoder is fused with combinatorial perturbation embeddings by cross-attention before predicting the perturbed profile $\hat{x}^{\mathrm{pert}}$.
\textbf{c}, \textit{ChemCellFlow} architecture for drug-response prediction, in which gene and chemical encoders define cell and perturbation states that are transformed by conditional optimal transport and normalizing-flow refinement before decoding $\hat{x}^{\mathrm{pert}}$.
\textbf{d}, Global and differential-expression prediction performance on the Adamson CRISPRi, Norman CRISPRa, and Srivatsan drug-response benchmarks. The upper row reports PCC and $-\log_{10}\mathrm{MSE}$ across all genes, whereas the lower row reports $\mathrm{PCC}_{\mathrm{DE}}$ and $-\log_{10}\mathrm{MSE}_{\mathrm{DE}}$ across observed differentially expressed genes. CellForge results are means $\pm$ s.d. across three independent runs.
\textbf{e}, ROC curves for recovery of the observed differentially expressed genes.
\textbf{f}, AUROC values for the recovery analysis in e.
\textbf{g}, Method rankings across global prediction, differential-expression prediction and DEG-recovery metrics; lower ranks indicate better performance.
\par\vspace{1.25em}

\refstepcounter{figure}
\label{fig:understudied_models}
\noindent\textbf{Fig. \thefigure{} \textbar{} CellForge models for perturbation prediction across modalities.}
\textbf{a}, \textit{totalGAT} architecture for paired RNA--protein CITE-seq data. Modality-specific encoders and cross-modal graph-attention layers integrate RNA features and protein markers, and separate decoders predict the perturbed RNA and protein profiles.
\textbf{b}, \textit{ChromDDPM} architecture for scATAC-seq data, using conditioning signals from cell state and perturbation context to guide denoising-based reconstruction of chromatin-accessibility profiles.
\textbf{c}, \textit{CPA-Traj} architecture for cytokine time-course data, in which a trajectory-aware variational autoencoder represents shared and cytokine-specific cell states that are combined with time and perturbation information before predicting expression changes.
\textbf{d}, Global and differential-feature prediction performance for Schiebinger RNA, Papalexi CITE-seq RNA and protein, and Liscovitch-Brauer ATAC readouts. The upper row shows PCC against $-\log_{10}\mathrm{MSE}$, and the lower row shows $\mathrm{PCC}_{\mathrm{DE}}$ against $-\log_{10}\mathrm{MSE}_{\mathrm{DE}}$. CellForge results are means $\pm$ s.d. across three independent runs.
\textbf{e}, ROC curves for recovery of the observed differential features.
\textbf{f}, AUROC values for the recovery analysis in e.
\textbf{g}, Method rankings across global prediction, differential-feature prediction and differential-feature recovery; lower ranks indicate better performance.
\par\vspace{1.25em}

\refstepcounter{figure}
\label{fig:llm-eval}
\noindent\textbf{Fig. \thefigure{} \textbar{} Blinded evaluation of CellForge-generated research plans.}
\textbf{a}, Example CellForge planning exchange for a single-cell perturbation task. Data-modeling, single-cell biology, and deep-learning experts propose complementary components. Peer review and critic feedback then guide the selected expert in revising the integrated plan before execution.
\textbf{b}, Blinded evaluation of plans generated by CellForge, OpenAI Deep Research, Perplexity Deep Research, Gemini Deep Research, Biomni, and a single-LLM baseline across drug, gene-knockout, and cytokine tasks. Five LLM judges scored each plan from 1 to 10 for scientific validity, technical feasibility, innovation, experimental design, and impact potential. Bars show mean scores, and error bars show variability across repeated evaluations.
\textbf{c}, Separate blinded human-expert evaluation of the same plans across technical novelty, feasibility, clarity and consistency, biological plausibility, mechanism explanation, pathway relevance, and cross-perturbation generalizability. Bars show mean scores across the three single-cell biology experts.
\textbf{d}, Inter-judge reliability among the five LLM judges, shown as pairwise Pearson correlations between scores. Off-diagonal correlations range from 0.88 to 0.93.
\textbf{e}, Associations of mean human-expert scores with CellForge's graph-based discussion confidence score for individual expert proposals (left) and with aggregate LLM-judge scores across plan-generating systems (right). Shapes denote task, and colors denote CellForge expert role (left) or plan-generating system (right).
\textbf{f}, Phase-resolved evaluation profiles for gene-knockout (left), drug-treatment (middle), and cytokine (right) tasks. In each radial plot, the upper sector reports method-design criteria and the lower sector reports task-analysis criteria. Lines denote plan-generating systems, and markers denote individual LLM judges.

\clearpage
\clearpage
\begin{center}
\includegraphics[width=\linewidth,height=0.95\textheight,keepaspectratio]{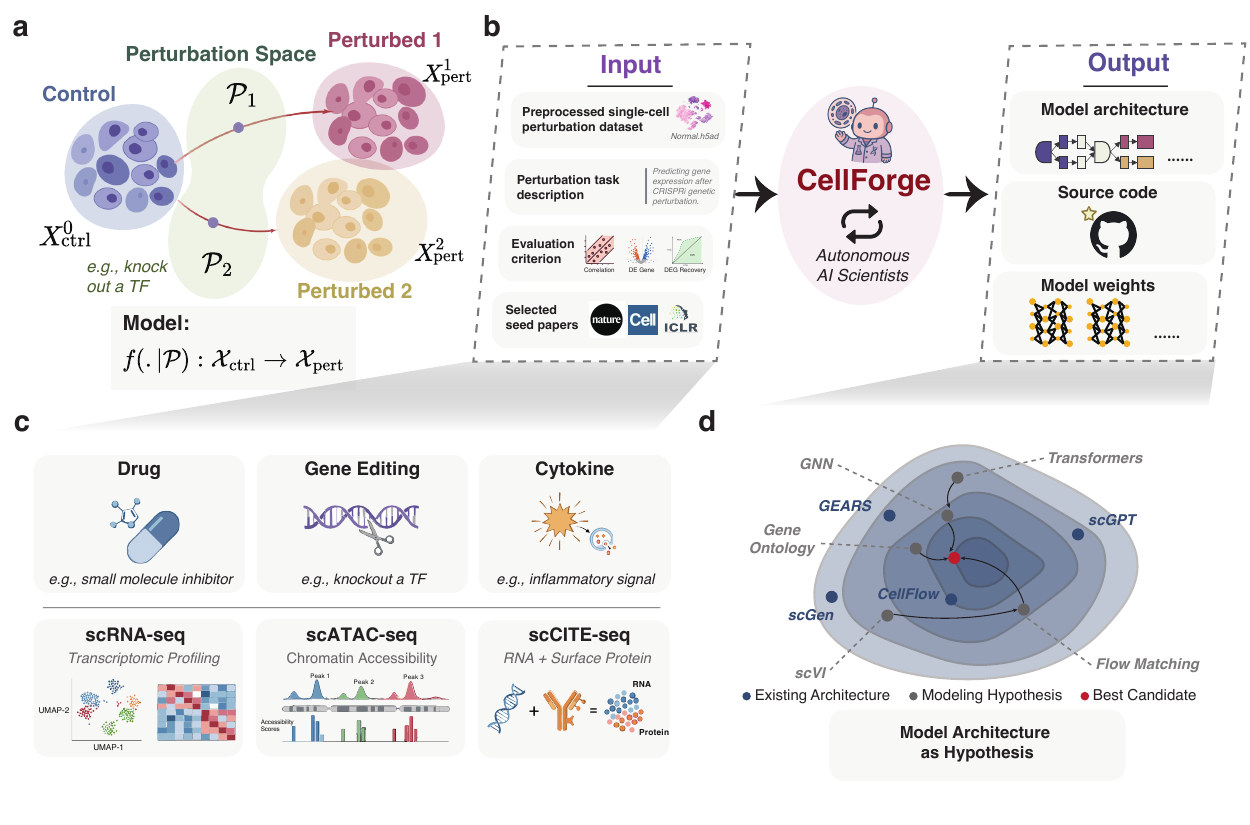}
\end{center}

\clearpage
\begin{center}
\includegraphics[width=\linewidth,height=0.95\textheight,keepaspectratio]{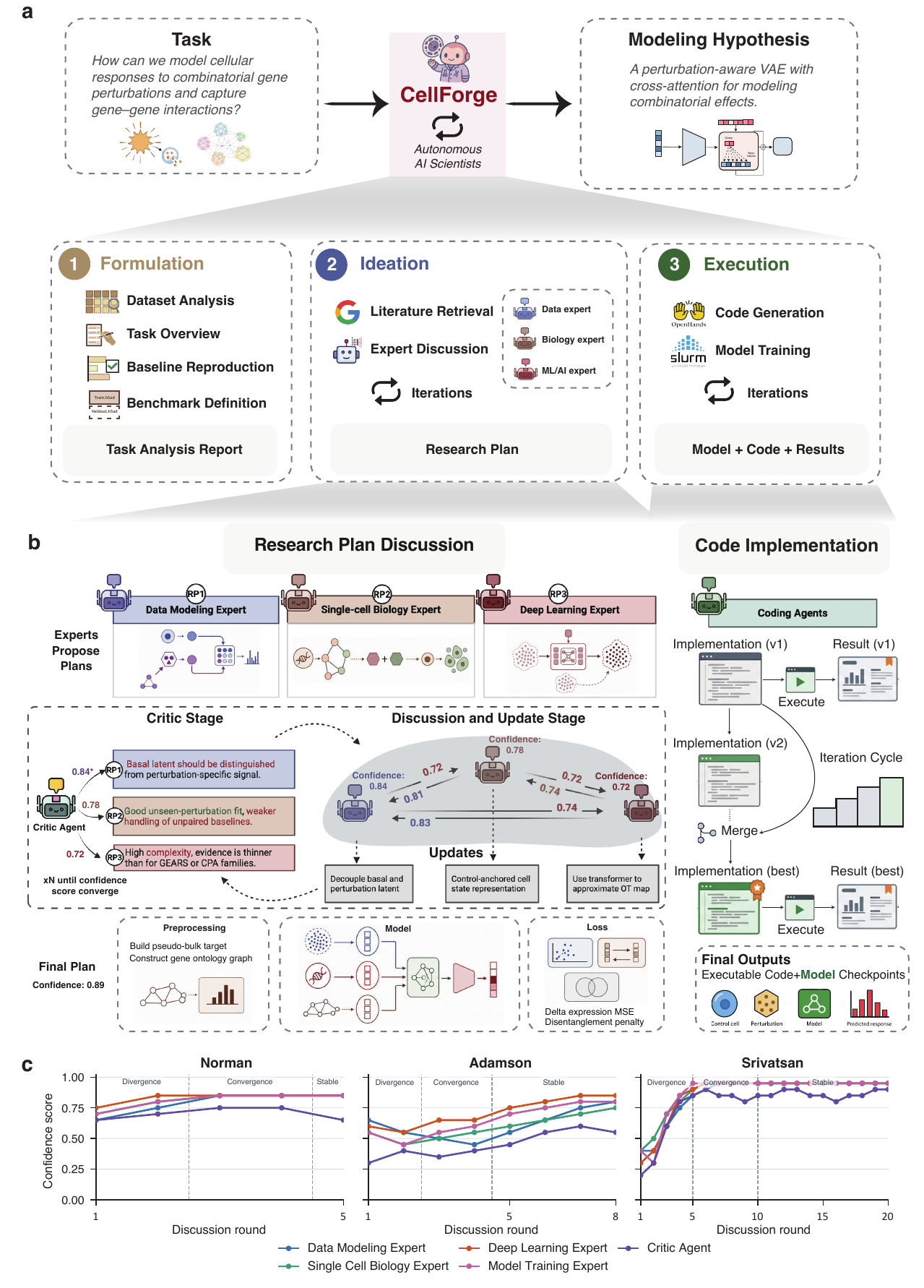}
\end{center}

\clearpage
\begin{center}
\includegraphics[width=\linewidth,height=0.95\textheight,keepaspectratio]{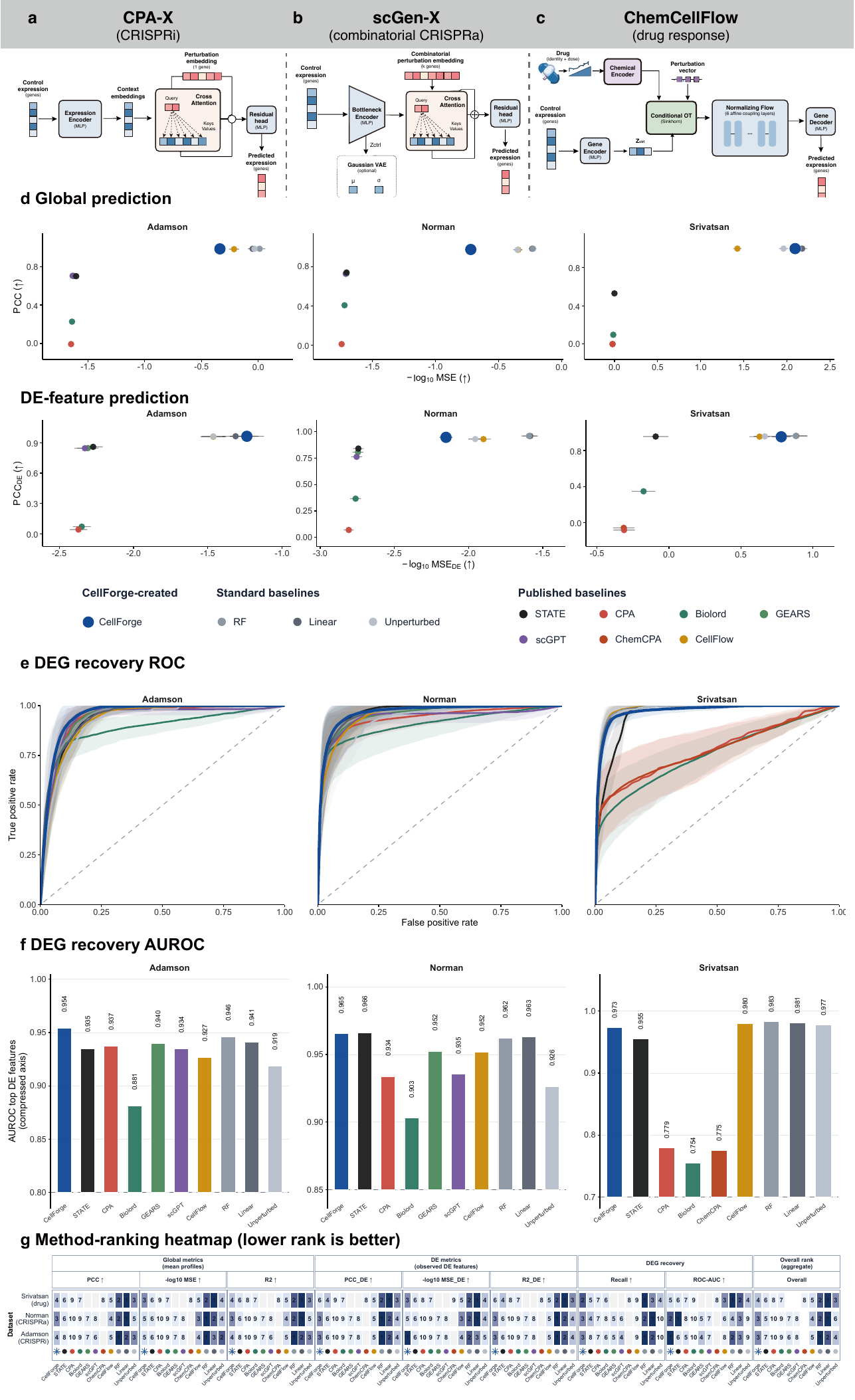}
\end{center}

\clearpage
\begin{center}
\includegraphics[width=\linewidth,height=0.9\textheight,keepaspectratio]{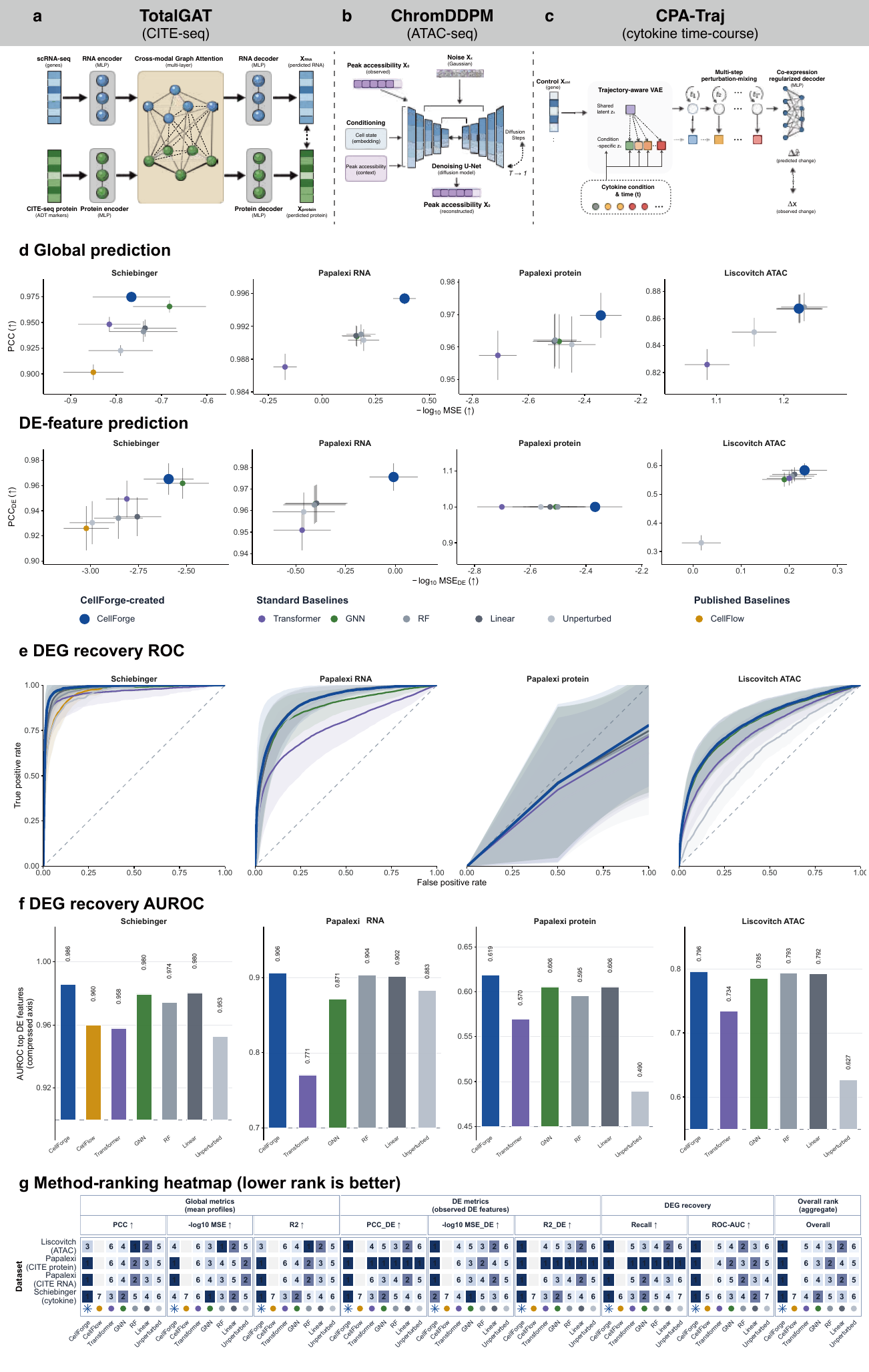}
\end{center}

\clearpage
\begin{center}
\includegraphics[width=\linewidth,height=0.95\textheight,keepaspectratio]{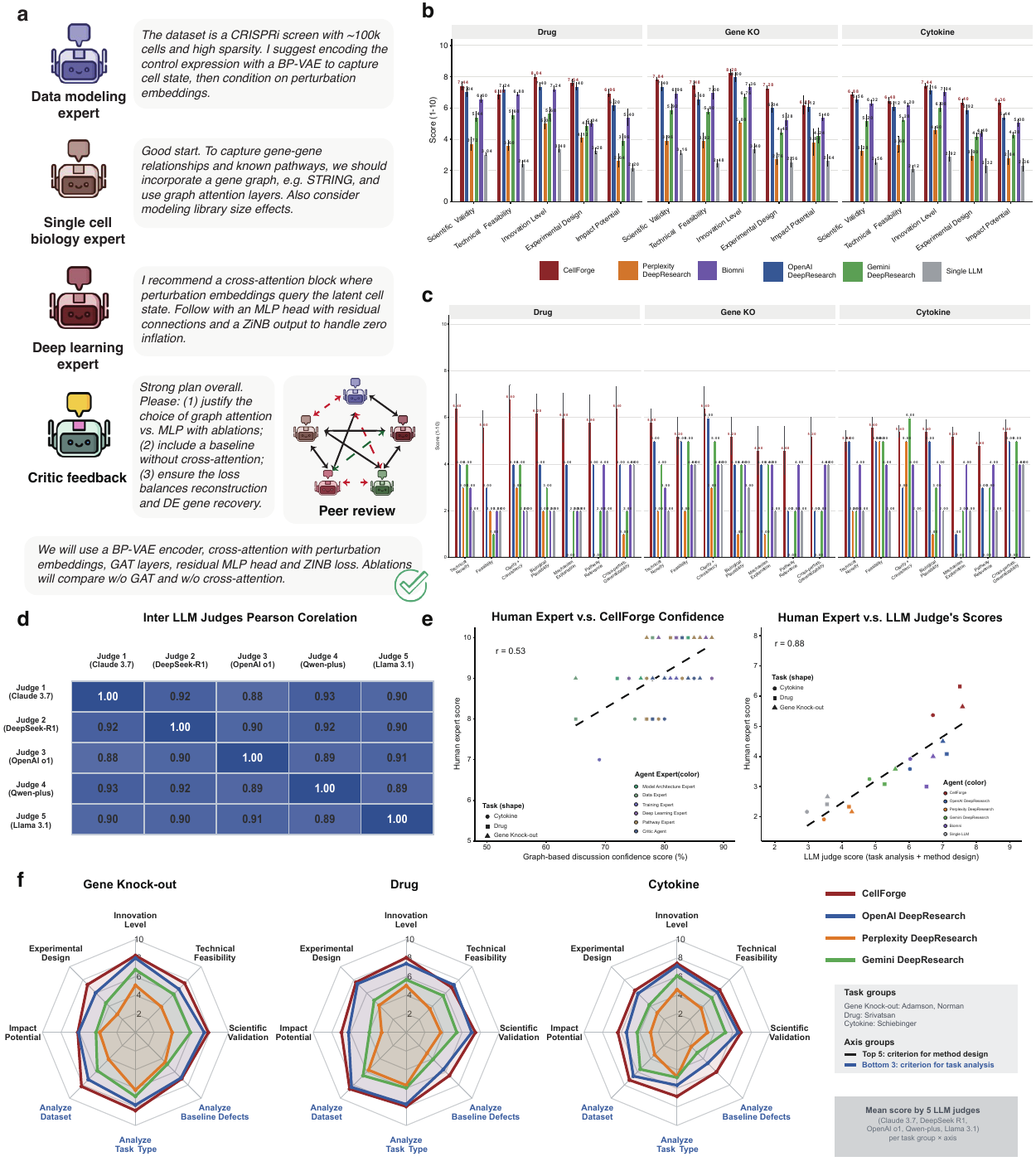}
\end{center}

\clearpage
\clearpage
\appendix

\section{Supplementary Results}
This section provides supporting analyses that complement the main results and clarifies how several claims in the main text should be interpreted.

\subsection{Representative Architecture Motifs}
\label{app:designed_models}

\paragraph{Analysis protocol.}
To examine how CellForge adapts model design to task structure, we ran the framework three times with different random seeds on each of the six benchmark datasets and inspected the executable training scripts produced by the Execution stage. The descriptions below correspond to the architectures that were actually committed to code, trained, and evaluated for each dataset. The most consistently recurring run-log label for each dataset is adopted as its canonical model name in the main text (\textit{CPA-X}, \textit{scGen-X}, \textit{ChemCellFlow}, \textit{CPA-Traj}, \textit{totalGAT}, \textit{ChromDDPM}); these names reflect the dominant generated configuration and should not be read as standardized or novel model classes.

\subsubsection*{Genetic Perturbation Settings}

\paragraph{Adamson CRISPRi --- \textit{CPA-X}.}
For single-gene CRISPRi, repeated runs converged on a feed-forward cross-attention design in which perturbation identity conditions a learned transformation of the control profile, related in spirit to scGen and CPA~\cite{lotfollahi2019scgen,lotfollahi2023predictinga}. The dominant configuration, \textit{CPA-X}, consists of an MLP expression encoder that maps the control profile $x_{\text{ctrl}}$ into a shared hidden space, a parallel MLP perturbation encoder for the one-hot perturbation vector, a multi-head cross-attention block in which the perturbation embedding attends to the encoded cell state, a residual MLP body, and a linear output head that predicts the post-perturbation expression profile. The model is trained with mean squared error on the perturbed profile against the observed target. Run-to-run variation was restricted in structural family: the main differences across runs were hidden width, number of attention heads, residual MLP depth, and dropout levels set by the hyperparameter search. Recurring structural motifs across runs included a single multi-head cross-attention operator between state and perturbation, a lightweight residual MLP body rather than a deep decoder stack, and absolute-profile regression rather than delta prediction.

\paragraph{Norman CRISPRa --- \textit{scGen-X}.}
The combinatorial CRISPRa setting, with its paired perturbations and stronger non-additive effects, pushed generated designs toward higher-capacity perturbation--state fusion rather than simple additive shifts, closer in motivation to GEARS and AttentionPert~\cite{roohani2024predicting,bai2024attentionpert}. The dominant configuration, \textit{scGen-X}, projects the high-dimensional control profile through a two-stage MLP bottleneck, optionally routes the bottleneck representation through a Gaussian VAE latent when toggled on, encodes perturbations through a linear perturbation embedding, and performs a single-token cross-attention in which the perturbation embedding queries the encoded expression state. The fused representation is refined by a residual MLP body and mapped to the predicted profile by a two-layer output head. Recurring structural motifs across runs included the expression bottleneck, the query-side perturbation cross-attention operator, and a hybrid selector that turned the VAE branch on or off depending on the run. Across seeds, combinatorial perturbation data consistently drove the design loop toward high-capacity attention-based fusion rather than simple latent shifts from a single perturbation token.

\subsubsection*{Drug Perturbation Setting}

\paragraph{Srivatsan sci-Plex --- \textit{ChemCellFlow}.}
The drug perturbation setting generated the widest range of proposals across seeds, consistent with the weaker and less stable benchmark performance. The recurrent motifs were nonetheless interpretable: most successful runs introduced an explicit dose-response component, an optimal-transport latent alignment step, and a flow-based refinement stage, making the resulting designs closer in spirit to CellFlow and neural optimal transport approaches~\cite{klein2025cellflow,bunne2023learning,hetzel2022predicting} than to generic transcriptomic autoencoders. The canonical configuration, \textit{ChemCellFlow}, combines a gene encoder that maps the control profile to a latent $z_{\text{ctrl}}$; a chemical encoder that maps perturbation identity and dose through separate subencoders and fuses them into a perturbation vector $p$; a Sinkhorn-based conditional optimal-transport block that computes a transport plan between control and perturbed latents; a six-layer normalizing-flow refiner composed of affine coupling layers and conditioned on $p$; and a modular gene decoder with ten gene-group output heads and a global residual path that maps the refined latent back to gene expression. The training loss combines reconstruction, optimal transport, flow consistency, and co-expression graph regularization terms. Recurring structural motifs across runs included the separate dose-encoder branch, the OT-plus-flow latent-refinement stack, and the modular gene-group decoder.

\subsubsection*{Temporal and Multi-modal Settings}

\paragraph{Schiebinger cytokine --- \textit{CPA-Traj}.}
On the cytokine reprogramming dataset, the recurring pattern was not simply a larger encoder-decoder, but explicit treatment of developmental trajectory: time-aware latent structure, perturbation-conditioned mixing, and co-expression-regularized decoding. The canonical configuration, \textit{CPA-Traj}, is built from three components. A trajectory-aware encoder embeds the control profile through an MLP and fuses it with a low-dimensional continuous time vector, then produces two parallel Gaussian latents (a shared trajectory latent and a condition-specific latent), each with its own $\mu$/$\log\sigma^2$ head, concatenated into a single latent $z$. A perturbation-mixing module embeds the one-hot perturbation linearly, concatenates it with $z$, and iteratively refines the mixture through three gated MLP blocks that repeatedly re-inject $z$ as a residual. A graph-regularized decoder maps the mixed latent to the predicted expression via a three-layer MLP and applies a learned linear projection whose output is regularized by a gene--gene co-expression matrix estimated during training. Recurring structural motifs across runs included the shared-versus-condition latent split, the multi-step mixing block with $z$-residuals, and the co-expression regularization on the decoder output. These designs were oriented toward capturing continuous developmental dynamics rather than predicting a single end-state distribution.

\paragraph{Papalexi CITE-seq --- \textit{totalGAT}.}
In the paired RNA--protein CITE-seq dataset, CellForge used \textit{totalGAT}, a joint architecture with modality-specific encoders, perturbation fusion, graph-attention layers, and transformer-style refinement applied to multi-token projections of the fused representation. The RNA branch maps the baseline expression profile through an MLP bottleneck, concatenates it with a perturbation embedding, reshapes the fused state into learned tokens, and predicts the perturbed RNA profile with an auxiliary perturbation-reconstruction head. The protein branch encodes the baseline protein profile through residual MLP blocks, projects it into eight synthetic node tokens, passes these tokens through graph-attention layers~\cite{velickovic2018gat}, combines the pooled protein representation with a perturbation embedding through bidirectional cross-attention, and predicts the perturbed protein profile with a residual MLP head. Both readouts are trained with MSE-based objectives against the observed profile. Recurring structural motifs across runs included modality-specific encoders, multi-token projection before attention-based refinement, perturbation-aware fusion, auxiliary perturbation reconstruction, and separate RNA and protein output heads.

\subsubsection*{Chromatin Accessibility Setting}

\paragraph{Liscovitch-Brauer scATAC-seq --- \textit{ChromDDPM}.}
The scATAC-seq dataset produced the sharpest shift away from standard scRNA-seq modeling assumptions, with generated designs moving toward denoising diffusion rather than direct regression. The canonical configuration, \textit{ChromDDPM}, inherits the DDPM formulation of Ho et al.~\cite{ho2020ddpm} applied to the sparse accessibility setting: it first maps the baseline accessibility vector through a two-stage MLP bottleneck, then conditions on a sinusoidal-time embedding of the diffusion step passed through a small MLP together with a linear perturbation embedding, concatenating both into a conditioning vector and projecting it to the hidden dimension. The noisy input and the conditioning vector are summed and passed through a four-block residual MLP, which produces a predicted noise vector and, through a parallel head, a predicted absolute accessibility profile; a secondary auxiliary head reconstructs the perturbation identity. The accompanying diffusion wrapper implements a standard DDPM-style forward process (linear $\beta$ schedule, cumulative $\bar{\alpha}$) with $q_{\text{sample}}$ and $p_{\text{sample}}$ routines over up to one thousand steps. Recurring structural motifs across runs included the sinusoidal-time-plus-perturbation conditioning vector, the residual MLP denoiser body, and the auxiliary perturbation-reconstruction head used to keep the denoiser perturbation-aware. Given the modest predictive performance in this setting, we interpret \textit{ChromDDPM} as a plausible adaptation of denoising diffusion to sparse regulatory data rather than a validated chromatin-specific architecture.

\subsection{Execution failure modes}
\label{app:failure_case_analysis}
We manually reviewed 20 failed-execution cases and grouped them into recurring categories. In this small-sample review, implementation and execution issues were the most common categories observed. Table~\ref{tab:failure_modes} summarizes this error profile and was used to guide lightweight safeguards during repair, especially explicit tensor-shape tracing and stricter validation of intermediate datatypes.

\begin{table}[t]
\centering
\small
\caption{Observed failure modes in a manual review of failed CellForge executions. Percentages sum to 100\%.}
\label{tab:failure_modes}
\begin{tabular}{lc}
\toprule
\textbf{Failure mode} & \textbf{Share of reviewed failures} \\
\midrule
Computation execution error & 41\% \\
Invalid type or unsupported operation & 23\% \\
Error recovery failure & 16\% \\
Model configuration error & 6\% \\
Data access failure & 5\% \\
Other system-level failures & 5\% \\
Hallucinated structures or outputs & 4\% \\
\bottomrule
\end{tabular}
\end{table}

\subsection{Compute profile and cost estimation}
\label{app:cost_analysis}
For reproducibility, we report hardware configuration and approximate token usage ranges as contextual compute information. These values are intended to convey order of magnitude rather than serve as an audited per-dataset cost analysis.

\paragraph{Cluster environment and scheduling.}
All CellForge-designed models reported in this paper, with parameter counts between 10 million and 30 million, were trained and evaluated through Slurm on Yale Center for Research Computing infrastructure, primarily on the \texttt{misha} and \texttt{bouchet} clusters. Production runs were submitted with \texttt{sbatch}. Each job requested one GPU (\texttt{--gpus=1}) together with CPU and memory resources and used a wall-time limit of \texttt{1-00:00:00}. Iterative refinement was implemented as a sequence of Slurm jobs, so each additional repair or retraining round consumed a new one-GPU batch allocation. Because the exact node assignment varied with scheduler allocation, we report the Slurm request pattern and runtime range rather than a single hardware SKU.

\paragraph{Cost interpretation.}
The multi-agent workflow incurs distinct LLM costs in task preparation, research ideation, and experiment execution. Table~\ref{tab:token_breakdown} reports internal development-time token-usage ranges intended to show scale rather than audited per-dataset totals. We do not report a single dollar figure because API cost depends on backend choice, service tier, and execution date. These estimates also exclude shared-cluster allocation costs, storage, and the minimal human safety checkpoints described in the main Methods. In internal runs, end-to-end execution typically completed within roughly 4--8 hours of wall-clock time once scheduled, with most of that time spent in model training rather than agent interaction; queue delay depended on partition load and is not included in this estimate.

\begin{table}[t]
\centering
\small
\caption{Illustrative token usage ranges per CellForge run, grouped by task complexity.}
\label{tab:token_breakdown}
\begin{tabular}{lcccc}
\toprule
\textbf{Phase} & \multicolumn{2}{c}{\textbf{Simple tasks}} & \multicolumn{2}{c}{\textbf{Complex tasks}} \\
 & Prompt & Completion & Prompt & Completion \\
\midrule
Task Preparation & 15{,}000 & 50{,}000 & 25{,}000 & 100{,}000 \\
Research Ideation & 20{,}000 & 100{,}000 & 40{,}000 & 200{,}000 \\
Experiment Execution & 5{,}000 & 50{,}000 & 15{,}000 & 100{,}000 \\
\midrule
\textbf{Total} & 40{,}000 & 200{,}000 & 80{,}000 & 400{,}000 \\
\bottomrule
\end{tabular}
\end{table}

\section{Supplementary Retrieval Corpus Inventory}
\label{app:knowledge_base}
The following lists summarize the fixed 46-paper corpus used by Agentic Retrieval. This local paper collection is stored in Qdrant so the agents can retrieve semantically related references before deciding whether to expand outward to PubMed or the web. For readability, we group papers by functional role in the retrieval stack: benchmark and review papers, perturbation-response models, multimodal integration methods, foundation models and language-model interfaces, and resource or tooling papers. Each entry reports the paper title, a shortened author line, venue information, and a one-sentence note on why the paper is useful during retrieval and synthesis.

\paragraph{Benchmark and review papers.}
\begin{enumerate}[label={[\arabic*]}, ref=\arabic*, leftmargin=0.7cm, itemsep=0em, topsep=0pt]
    \small
    \item \href{https://doi.org/10.1186/s13059-019-1795-z}{A comparison of automatic cell identification methods for single-cell RNA sequencing data}. \textbf{Genome Biology (2019).}
    \newline
    Tamim Abdelaal, Lieke Michielsen, \ldots, Marcel J. T. Reinders, Ahmed Mahfouz.
    \newline
    \emph{This benchmark systematically compares automated cell-type annotation methods for scRNA-seq, giving CellForge a practical reference for how downstream prediction pipelines should interpret and evaluate cell-state labels.}
    \item \href{https://doi.org/10.1016/j.csbj.2024.04.058}{A mini-review on perturbation modelling across single-cell omic modalities}. \textbf{Computational and Structural Biotechnology Journal (2024).}
    \newline
    George I. Gavriilidis, Vasileios Vasileiou, \ldots, Naveed Ishaque, Fotis Psomopoulos.
    \newline
    \emph{This review organizes the landscape of single-cell perturbation modeling across modalities, making it useful for retrieval when the agent needs a compact map of task settings, data types, and modeling choices.}
    \item \href{https://www.nature.com/articles/s41592-021-01336-8}{Benchmarking atlas-level data integration in single-cell genomics}. \textbf{Nature Methods (2022).}
    \newline
    Malte D. Luecken, M. B{\"u}ttner, \ldots, M. {Colom{\'e}-Tatch{\'e}}, Fabian J. Theis.
    \newline
    \emph{This paper benchmarks large-scale single-cell integration methods and evaluation metrics, which helps CellForge reason about batch correction, biological conservation, and the tradeoffs of joint embedding strategies.}
    \item \href{http://arxiv.org/abs/2410.13956}{Benchmarking Transcriptomics Foundation Models for Perturbation Analysis: one PCA still rules them all}. \textbf{arXiv (2024).}
    \newline
    Ihab Bendidi, Shawn Whitfield, \ldots, Emmanuel Noutahi, Alisandra K. Denton.
    \newline
    \emph{This study shows that strong linear baselines can still outperform many transcriptomics foundation models on perturbation tasks, providing a useful caution against overcommitting retrieval to foundation-model-heavy solutions.}
    \item \href{https://www.nature.com/articles/s41576-023-00586-w}{Best Practices for Single-Cell Analysis across Modalities}. \textbf{Nature Reviews Genetics (2023).}
    \newline
    Lukas Heumos, Anna C. Schaar, \ldots, Herbert B. Schiller, Fabian J. Theis.
    \newline
    \emph{This best-practices review summarizes modality-aware preprocessing, quality control, integration, and interpretation decisions, so it serves as a methodological anchor for robust pipeline construction.}
    \item \href{https://doi.org/10.1093/bib/bbab531}{Deep Learning Tackles Single-Cell Analysis-a Survey of Deep Learning for scRNA-seq Analysis}. \textbf{Briefings in Bioinformatics (2022).}
    \newline
    Mario Flores, Zhentao Liu, \ldots, Yidong Chen, Yufei Huang.
    \newline
    \emph{This survey distills the main deep-learning families used in scRNA-seq analysis, helping retrieval connect a new task to the relevant representation-learning, generative, and predictive modeling paradigms.}
    \item \href{https://www.nature.com/articles/s41587-025-02694-w}{Defining and Benchmarking Open Problems in Single-Cell Analysis}. \textbf{Nature Biotechnology (2025).}
    \newline
    Malte D. Luecken, Scott Gigante, \ldots, Fabian J. Theis, Smita Krishnaswamy.
    \newline
    \emph{This paper formalizes open benchmark tasks and evaluation principles for single-cell analysis, which is directly useful when CellForge needs to frame a prediction problem and choose defensible metrics.}
    \item \href{https://www.cell.com/cell-systems/abstract/S2405-4712(21)00202-7}{Machine Learning for Perturbational Single-Cell Omics}. \textbf{Cell Systems (2021).}
    \newline
    Yuge Ji, Mohammad Lotfollahi, F. Alexander Wolf, Fabian J. Theis.
    \newline
    \emph{This review lays out the core prediction tasks, datasets, and model classes in perturbational single-cell omics, making it a central retrieval paper for orienting architecture design around the actual problem taxonomy.}
    \item \href{https://www.sciencedirect.com/science/article/pii/S2001037025000583}{Machine Learning to Dissect Perturbations in Complex Cellular Systems}. \textbf{Computational and Structural Biotechnology Journal (2025).}
    \newline
    Pablo {Monfort-Lanzas}, Katja Rungger, Leonie Madersbacher, Hubert Hackl.
    \newline
    \emph{This review expands perturbation modeling beyond simple single-cell settings to broader cellular systems, helping retrieval connect CellForge to methods that capture complex responses, interactions, and experimental contexts.}
    \item \href{https://www.nature.com/articles/s41587-020-00803-5}{Quantifying the Effect of Experimental Perturbations at Single-Cell Resolution}. \textbf{Nature Biotechnology (2021).}
    \newline
    Daniel B. Burkhardt, Jay S. Stanley, \ldots, David {van Dijk}, Smita Krishnaswamy.
    \newline
    \emph{This work develops principled ways to measure perturbation effects at single-cell resolution, which is valuable for retrieval when the agent must reason about effect size, response structure, and evaluation targets.}
    \item \href{https://www.nature.com/articles/s42256-024-00948-x}{Reply to: Deeper Evaluation of a Single-Cell Foundation Model}. \textbf{Nature Machine Intelligence (2024).}
    \newline
    Fan Yang, Fang Wang, \ldots, Junzhou Huang, Jianhua Yao.
    \newline
    \emph{This reply sharpens the debate over how single-cell foundation models should be evaluated, making it useful for retrieval whenever CellForge needs to weigh claimed generalization against benchmark design and critique.}
    \item \href{https://genomebiology.biomedcentral.com/articles/10.1186/s13059-024-03446-z}{Zero-Shot Evaluation Reveals Limitations of Single-Cell Foundation Models}. \textbf{Genome Biology (2025).}
    \newline
    Kasia Z. Kedzierska, Lorin Crawford, Ava P. Amini, Alex X. Lu.
    \newline
    \emph{This paper documents substantial limits of zero-shot single-cell foundation models on benchmark tasks, giving CellForge concrete evidence that pretrained embeddings alone may not be sufficient for perturbation prediction.}
\end{enumerate}

\paragraph{Perturbation-response models.}
\begin{enumerate}[label={[\arabic*]}, ref=\arabic*, leftmargin=0.7cm, itemsep=0em, topsep=0pt, start=13]
    \small
    \item \href{https://www.nature.com/articles/s41587-020-0605-1}{Cell Type Prioritization in Single-Cell Data}. \textbf{Nature Biotechnology (2021).}
    \newline
    Michael A. Skinnider, Jordan W. Squair, \ldots, Ariel J. Levine, Gr{\'e}goire Courtine.
    \newline
    \emph{This paper turns perturbation effects into a cell-type ranking problem, giving CellForge a concrete example of how transcriptional signatures can be used to prioritize responsive populations rather than only predict average expression shifts.}
    \item \href{https://www.cell.com/cell-systems/abstract/S2405-4712(20)30464-6}{CellBox: Interpretable Machine Learning for Perturbation Biology with Application to the Design of Cancer Combination Therapy}. \textbf{Cell Systems (2021).}
    \newline
    Bo Yuan, Ciyue Shen, \ldots, John Ingraham, Chris Sander.
    \newline
    \emph{CellBox shows that dynamic, interpretable models can simulate multi-perturbation responses and support intervention design, which is directly relevant when CellForge retrieves mechanistic alternatives to purely black-box predictors.}
    \item \href{https://www.nature.com/articles/s41556-025-01626-9}{Decoding Heterogeneous Single-Cell Perturbation Responses}. \textbf{Nature Cell Biology (2025).}
    \newline
    Bicna Song, Dingyu Liu, \ldots, Danwei Huangfu, Wei Li.
    \newline
    \emph{This work emphasizes that perturbation effects are strongly cell-state dependent and heterogeneous across cells, making it useful when CellForge needs literature support for designs that model subpopulation-specific responses instead of a single pooled effect.}
    \item \href{https://www.nature.com/articles/s41586-022-05688-9}{Dissecting Cell Identity via Network Inference and in Silico Gene Perturbation}. \textbf{Nature (2023).}
    \newline
    Kenji Kamimoto, Blerta Stringa, \ldots, Lilianna {Solnica-Krezel}, Samantha A. Morris.
    \newline
    \emph{This paper infers regulatory networks and uses in silico perturbation to probe cell identity, offering CellForge a mechanistic template for retrieval when candidate models should incorporate gene-regulatory structure rather than rely only on direct supervised prediction.}
    \item \href{https://openreview.net/forum?id=Tmx4o3Jg55}{LangPert: LLM-Driven Contextual Synthesis for Unseen Perturbation Prediction}. \textbf{ICLR 2025 Workshop on Machine Learning for Genomics Explorations.}
    \newline
    Kaspar M{\"a}rtens, Marc Boubnovski Martell, Cesar A. {Prada-Medina}, Rory {Donovan-Maiye}.
    \newline
    \emph{LangPert is especially relevant to CellForge because it uses language-model-based contextual synthesis for unseen perturbation prediction, directly overlapping with the idea of using retrieved knowledge to propose models beyond the observed perturbation space.}
    \item \href{https://www.embopress.org/doi/full/10.1038/s44320-025-00131-3}{PerturbNet Predicts Single-Cell Responses to Unseen Chemical and Genetic Perturbations}. \textbf{Molecular Systems Biology (2025).}
    \newline
    Hengshi Yu, Weizhou Qian, Yuxuan Song, Joshua D Welch.
    \newline
    \emph{PerturbNet targets generalization to unseen chemical and genetic interventions, making it a strong retrieval anchor for CellForge when the task requires architectures that extrapolate beyond perturbations present during training.}
    \item \href{https://www.embopress.org/doi/full/10.15252/msb.202211517}{Predicting Cellular Responses to Complex Perturbations in High-throughput Screens}. \textbf{Molecular Systems Biology (2023).}
    \newline
    Mohammad Lotfollahi, Anna Klimovskaia Susmelj, \ldots, David Lopez-Paz, Fabian J Theis.
    \newline
    \emph{This paper addresses compositional and high-order perturbation settings in large screening datasets, which is useful for CellForge when retrieved evidence needs to justify models for combinations rather than single interventions.}
    \item \href{https://www.nature.com/articles/s41587-023-01905-6}{Predicting Transcriptional Outcomes of Novel Multigene Perturbations with GEARS}. \textbf{Nature Biotechnology (2024).}
    \newline
    Yusuf Roohani, Kexin Huang, Jure Leskovec.
    \newline
    \emph{GEARS demonstrates that gene-interaction structure can improve prediction for novel multigene perturbations, giving CellForge a concrete precedent for graph-aware architectures in combinatorial CRISPR settings.}
    \item \href{https://www.nature.com/articles/s41467-024-53457-1}{Predicting Transcriptional Responses to Novel Chemical Perturbations Using Deep Generative Model for Drug Discovery}. \textbf{Nature Communications (2024).}
    \newline
    Xiaoning Qi, Lianhe Zhao, \ldots, Shengyong Yang, Yi Zhao.
    \newline
    \emph{This paper extends deep generative perturbation modeling to novel chemical compounds, which helps CellForge retrieve evidence for architectures that bridge single-cell response prediction with drug-discovery-oriented chemical generalization.}
    \item \href{https://www.nature.com/articles/s41592-019-0494-8}{scGen Predicts Single-Cell Perturbation Responses}. \textbf{Nature Methods (2019).}
    \newline
    Mohammad Lotfollahi, F. Alexander Wolf, Fabian J. Theis.
    \newline
    \emph{scGen is a foundational perturbation-prediction baseline that models responses as latent-space shifts in a conditional variational framework, so it remains an essential reference point for CellForge when comparing against standard generative approaches.}
    \item \href{https://www.nature.com/articles/s41467-021-22197-x}{scGNN Is a Novel Graph Neural Network Framework for Single-Cell RNA-Seq Analyses}. \textbf{Nature Communications (2021).}
    \newline
    Juexin Wang, Anjun Ma, \ldots, Qin Ma, Dong Xu.
    \newline
    \emph{Although broader than perturbation prediction alone, scGNN shows how graph neural networks can capture cell-cell and gene-gene structure in single-cell data, making it relevant when CellForge retrieves graph-based building blocks for perturbation models.}
    \item \href{https://papers.nips.cc/paper/2022}{Supervised training of conditional monge maps}. \textbf{Advances in Neural Information Processing Systems (2022).}
    \newline
    Charlotte Bunne, Andreas Krause, Marco Cuturi.
    \newline
    \emph{This paper provides an optimal-transport formulation for learning conditional mappings between cellular states, which is valuable for CellForge retrieval when transport-based approaches are better suited than latent VAEs for modeling perturbation-induced state transitions.}
\end{enumerate}

\paragraph{Multimodal integration methods.}
\begin{enumerate}[label={[\arabic*]}, ref=\arabic*, leftmargin=0.7cm, itemsep=0em, topsep=0pt, start=25]
    \small
    \item \href{https://www.nature.com/articles/s41587-023-02097-9}{Characterizing the Impacts of Dataset Imbalance on Single-Cell Data Integration}. \textbf{Nature Biotechnology (2024).}
    \newline
    Hassaan Maan, Lin Zhang, \ldots, Kieran R. Campbell, Bo Wang.
    \newline
    \emph{This paper shows how batch-size and cell-type imbalance can distort integration performance, which helps CellForge reason about when integration errors reflect data skew rather than model quality.}
    \item \href{https://www.cell.com/cell/abstract/S0092-8674(21)00583-3}{Integrated Analysis of Multimodal Single-Cell Data}. \textbf{Cell (2021).}
    \newline
    Yuhan Hao, Stephanie Hao, \ldots, Peter Smibert, Rahul Satija.
    \newline
    \emph{This work introduces weighted nearest neighbors for joint RNA--protein and other multimodal analyses, making it a core reference for CITE-seq-style representation alignment.}
    \item \href{https://www.nature.com/articles/s41576-019-0093-7}{Integrative Single-Cell Analysis}. \textbf{Nature Reviews Genetics (2019).}
    \newline
    Tim Stuart, Rahul Satija.
    \newline
    \emph{This review organizes the main computational strategies for batch correction, reference mapping, and cross-dataset integration, giving CellForge a high-level map of the integration design space.}
    \item \href{https://www.nature.com/articles/s41592-020-01050-x}{Joint Probabilistic Modeling of Single-Cell Multi-Omic Data with totalVI}. \textbf{Nature Methods (2021).}
    \newline
    Adam Gayoso, Zo{\"e} Steier, \ldots, Aaron Streets, Nir Yosef.
    \newline
    \emph{This paper presents a joint probabilistic model for paired RNA and protein measurements that denoises both modalities in a shared latent space, making it directly relevant for multimodal perturbation modeling.}
    \item \href{https://www.nature.com/articles/s41592-023-01909-9}{MultiVI: Deep Generative Model for the Integration of Multimodal Data}. \textbf{Nature Methods (2023).}
    \newline
    Tal Ashuach, Mariano I. Gabitto, \ldots, Michael I. Jordan, Nir Yosef.
    \newline
    \emph{This work extends deep generative integration to RNA--ATAC data with partially missing modalities, which is useful when CellForge must handle sparse or incompletely paired multi-omic measurements.}
    \item \href{https://proceedings.neurips.cc/paper_files/paper/2023/hash/4db8a681ae1e58376dc6227978829063-Abstract-Conference.html}{MuSe-GNN: Learning Unified Gene Representation From Multimodal Biological Graph Data}. \textbf{NeurIPS (2023).}
    \newline
    Tianyu Liu, Yuge Wang, Rex Ying, Hongyu Zhao.
    \newline
    \emph{This paper learns unified gene embeddings from multimodal biological graphs, offering a graph-based route for incorporating structured biological priors into multimodal prediction.}
    \item \href{http://arxiv.org/abs/2303.00233}{Single-Cell Multimodal Prediction via Transformers}. \textbf{CIKM (2023).}
    \newline
    Wenzhuo Tang, Hongzhi Wen, \ldots, Hui Liu, Jiliang Tang.
    \newline
    \emph{This work applies transformer architectures to single-cell multimodal prediction, making it a useful reference for sequence-style fusion of cross-modal cellular features.}
\end{enumerate}

\paragraph{Foundation models and language-model interfaces.}
\begin{enumerate}[label={[\arabic*]}, ref=\arabic*, leftmargin=0.7cm, itemsep=0em, topsep=0pt, start=32]
    \small
    \item \href{https://proceedings.mlr.press/v235/levine24a.html}{Cell2Sentence: Teaching Large Language Models the Language of Biology}. \textbf{ICML (2024).}
    \newline
    Daniel Levine, Syed A. Rizvi, \ldots, Rahul Madhav Dhodapkar, David Van Dijk.
    \newline
    \emph{This paper maps single-cell states into natural-language-like representations, making it a useful reference for how LLM interfaces can absorb cellular information without discarding biological structure.}
    \item \href{https://doi.org/10.1093/bioinformatics/btab083}{DNABERT: pre-trained Bidirectional Encoder Representations from Transformers model for DNA-language in genome}. \textbf{Bioinformatics (2021).}
    \newline
    Yanrong Ji, Zhihan Zhou, Han Liu, Ramana V. Davuluri.
    \newline
    \emph{DNABERT established that transformer pretraining on genomic sequences learns transferable nucleotide representations, which motivates sequence-native foundation modeling in the CellForge retrieval set.}
    \item \href{https://www.nature.com/articles/s41422-024-01034-y}{GeneCompass: Deciphering Universal Gene Regulatory Mechanisms with a Knowledge-Informed X-Species Foundation Model}. \textbf{Cell Research (2024).}
    \newline
    Xiaodong Yang, Guole Liu, \ldots, Ge Yang, Xin Li.
    \newline
    \emph{GeneCompass combines large-scale pretraining with explicit biological knowledge to recover regulatory mechanisms across species, offering a concrete example of foundation models that encode gene regulation rather than only expression similarity.}
    \item \href{https://doi.org/10.1093/bioinformatics/btae075}{GeneGPT: Augmenting Large Language Models with Domain Tools for Improved Access to Biomedical Information}. \textbf{Bioinformatics (2024).}
    \newline
    Qiao Jin, Yifan Yang, Qingyu Chen, Zhiyong Lu.
    \newline
    \emph{GeneGPT shows how an LLM can call biomedical tools and databases during reasoning, which directly informs CellForge's use of retrieval-augmented agents instead of a closed-book language model.}
    \item \href{https://doi.org/10.1177/10943420231201154}{GenSLMs: Genome-scale Language Models Reveal SARS-CoV-2 Evolutionary Dynamics}. \textbf{The International Journal of High Performance Computing Applications (2023).}
    \newline
    Maxim Zvyagin, Alexander Brace, \ldots, Venkatram Vishwanath, Arvind Ramanathan.
    \newline
    \emph{GenSLMs demonstrates that genome-scale language models can capture evolutionary regularities from massive sequence corpora, broadening the retrieval pool beyond transcriptomics to biologically meaningful foundation-model pretraining at scale.}
    \item \href{https://www.biorxiv.org/content/10.1101/2024.10.23.619972v1}{scGenePT: Is Language All You Need for Modeling Single-Cell Perturbations?}. \textbf{bioRxiv (2024).}
    \newline
    Ana-Maria Istrate, Donghui Li, Theofanis Karaletsos.
    \newline
    \emph{scGenePT tests whether language-model-derived gene embeddings are sufficient for single-cell perturbation prediction, making it directly relevant to whether LLM-style priors help on CellForge's core task.}
    \item \href{https://www.nature.com/articles/s41592-024-02201-0}{scGPT: Toward Building a Foundation Model for Single-Cell Multi-Omics Using Generative AI}. \textbf{Nature Methods (2024).}
    \newline
    Haotian Cui, Chloe Wang, \ldots, Nan Duan, Bo Wang.
    \newline
    \emph{scGPT is a central reference for pretrained transformer modeling of single-cell multi-omics, providing a concrete baseline for how foundation models can represent cells, genes, and perturbational context in one framework.}
    \item \href{https://www.nature.com/articles/s41551-024-01284-6}{Simple and Effective Embedding Model for Single-Cell Biology Built from ChatGPT}. \textbf{Nature Biomedical Engineering (2025).}
    \newline
    Yiqun Chen, James Zou.
    \newline
    \emph{This work shows that strong single-cell embeddings can be distilled from ChatGPT-derived supervision with a relatively simple pipeline, which is useful as a counterpoint to heavier bespoke foundation-model architectures.}
    \item \href{https://arxiv.org/abs/2401.06199}{xTrimoPGLM: unified 100B-scale pre-trained transformer for deciphering the language of protein}. \textbf{arXiv (2024).}
    \newline
    Bo Chen, Xingyi Cheng, \ldots, Jie Tang, Le Song.
    \newline
    \emph{xTrimoPGLM extends large-scale biological language modeling to proteins at 100B-parameter scale, giving CellForge a retrieval anchor for foundation-model design patterns in molecular domains beyond transcriptomics.}
\end{enumerate}

\paragraph{Resources and tooling papers.}
\begin{enumerate}[label={[\arabic*]}, ref=\arabic*, leftmargin=0.7cm, itemsep=0em, topsep=0pt, start=41]
    \small
    \item \href{https://www.nature.com/articles/s41587-020-0470-y}{Combinatorial Single-Cell CRISPR Screens by Direct Guide RNA Capture and Targeted Sequencing}. \textbf{Nature Biotechnology (2020).}
    \newline
    Joseph M. Replogle, Thomas M. Norman, \ldots, Jonathan S. Weissman, Britt Adamson.
    \newline
    \emph{This paper establishes a scalable protocol for directly capturing guide identities in combinatorial Perturb-seq experiments, which matters because many CellForge tasks depend on precisely annotated perturbation labels.}
    \item \href{https://www.nature.com/articles/nprot.2017.016}{Genome-Scale CRISPR-Cas9 Knockout and Transcriptional Activation Screening}. \textbf{Nature Protocols (2017).}
    \newline
    Julia Joung, Silvana Konermann, \ldots, Neville E. Sanjana, Feng Zhang.
    \newline
    \emph{This protocol paper details practical genome-scale CRISPR knockout and activation screening workflows, grounding retrieval in the experimental setup behind CRISPRi and CRISPRa datasets.}
    \item \href{https://www.science.org/doi/full/10.1126/science.aax6234}{Massively Multiplex Chemical Transcriptomics at Single-Cell Resolution}. \textbf{Science (2020).}
    \newline
    Sanjay R. Srivatsan, Jos{\'e} L. {McFaline-Figueroa}, \ldots, Jay Shendure, Cole Trapnell.
    \newline
    \emph{This work introduces large-scale multiplexed chemical perturbation profiling at single-cell resolution, providing a canonical resource for drug-response prediction settings.}
    \item \href{https://genomebiology.biomedcentral.com/articles/10.1186/s13059-017-1382-0}{SCANPY: Large-Scale Single-Cell Gene Expression Data Analysis}. \textbf{Genome Biology (2018).}
    \newline
    F. Alexander Wolf, Philipp Angerer, Fabian J. Theis.
    \newline
    \emph{This paper provides the standard scalable analysis toolkit for preprocessing, neighborhood graphs, clustering, and visualization, so it anchors many downstream single-cell pipelines used around CellForge.}
    \item \href{https://www.biorxiv.org/content/10.1101/2025.02.27.640494v1}{scBaseCamp: An AI Agent-Curated, Uniformly Processed, and Continually Expanding Single Cell Data Repository}. \textbf{bioRxiv (2025).}
    \newline
    Nicholas D. Youngblut, Christopher Carpenter, \ldots, Hani Goodarzi, Yusuf H. Roohani.
    \newline
    \emph{This repository paper describes an agent-curated and uniformly processed single-cell data collection, which is relevant as a scalable source of standardized training and benchmarking datasets.}
    \item \href{https://www.nature.com/articles/s41592-023-02144-y}{scPerturb: Harmonized Single-Cell Perturbation Data}. \textbf{Nature Methods (2024).}
    \newline
    Stefan Peidli, Tessa D. Green, \ldots, Nils Bl{\"u}thgen, Chris Sander.
    \newline
    \emph{This work harmonizes diverse single-cell perturbation datasets into a common resource, making it especially valuable for retrieval over comparable benchmarks and evaluation settings.}
\end{enumerate}

\section{Prompt templates and example outputs}

This section collects the full prompt templates and representative output artifacts for the three CellForge team stages (Formulation, Ideation, and Execution), previously distributed across three separate team sections.

\subsection{Task-preparation prompt structure}
\label{app:taskprep_prompts}
\label{app:data_parser}
The Task Preparation team operates from three role-specific prompts: a Dataset Analyst for data characterisation, a Problem Investigator for prediction-task formulation, and a Baseline Assessor for reproducible reference benchmarks. Together they produce the task report that grounds the subsequent Ideation stage.

\begin{promptbox}{Dataset Analyst}
\textit{``You are an expert in single-cell dataset analysis, specializing in evaluating single-cell perturbation data. Your task is to provide a comprehensive analysis of dataset characteristics and quality, focusing on experimental design, data properties, preprocessing recommendations, and quality assessment.''}\newline
\{decision\_context\}\newline
\newline
\{design\_context\}\newline
\newline
1.~Analyze Experimental Design:\newline
- Evaluate experimental setup and biological conditions\newline
- Assess sample size and biological replicates\newline
- Review perturbation design and controls\newline
- Analyze biological variability and batch effects\newline
- Consider technical and biological validation\newline
\newline
2.~Assess Data Properties:\newline
- Analyze data distribution and biological patterns\newline
- Evaluate feature space and biological relevance\newline
- Assess data quality and technical artifacts\newline
- Identify biological and technical noise\newline
- Consider data sparsity and missing values\newline
\newline
3.~Provide Preprocessing Recommendations:\newline
- Suggest normalization strategies for biological data\newline
- Recommend quality control procedures\newline
- Propose feature selection methods\newline
- Design batch effect correction approaches\newline
- Plan biological validation steps\newline
\newline
4.~Conduct Quality Assessment:\newline
- Evaluate data completeness and coverage\newline
- Assess biological relevance and interpretability\newline
- Analyze technical quality and reproducibility\newline
- Consider biological validation requirements\newline
- Identify potential limitations and biases\newline
\newline
Task Description:\newline
\{task\_description\}\newline
\newline
Dataset Information:\newline
\{json.dumps(dataset\_info, indent=2)\}\newline
\newline
Relevant Literature:\newline
\{papers\_context\}\newline
\newline
Please provide a comprehensive analysis in the following JSON format, incorporating decision support information:
\end{promptbox}

\begin{promptbox}{Problem Investigator}
\textit{``You are an expert in biological research problem analysis with extensive experience in designing computational solutions for biological applications. Your task is to provide a comprehensive investigation of the research problem and design a solution approach, focusing on problem definition, key challenges, research questions, and analysis methods.''}\newline
1.~Define Research Problem:\newline
- Formally define the problem in biological context\newline
- Identify key biological variables and their relationships\newline
- Specify input-output mappings and biological constraints\newline
- Define evaluation criteria with biological significance\newline
- Establish success metrics with biological validation\newline
\newline
2.~Analyze Key Challenges:\newline
- Identify biological and technical challenges\newline
- Assess data quality and biological variability\newline
- Evaluate computational complexity and scalability\newline
- Consider biological interpretability requirements\newline
- Address validation and reproducibility concerns\newline
\newline
3.~Formulate Research Questions:\newline
- Define primary research questions with biological focus\newline
- Identify key hypotheses to be tested\newline
- Specify biological mechanisms to be investigated\newline
- Outline experimental validation requirements\newline
- Establish biological significance criteria\newline
\newline
4.~Design Analysis Methods:\newline
- Propose computational approaches with biological relevance\newline
- Design validation strategies with biological context\newline
- Specify analysis pipelines with biological interpretation\newline
- Plan experimental validation with biological controls\newline
- Establish reproducibility standards with biological validation
\end{promptbox}

\begin{promptbox}{Baseline Assessor}
\textit{``You are an expert in deep learning model evaluation and benchmarking for single cell perturbation prediction, with extensive experience in assessing machine learning models for biological applications. Your task is to provide a comprehensive assessment of baseline models and evaluation strategies, focusing on model analysis, evaluation framework, performance analysis, and improvement suggestions.''}\newline
\{framework\_context\}\newline
\newline
\{guide\_context\}\newline
\newline
Dynamic Analysis of Existing Methods (auto-extracted from retrieved literature):\newline
\{method\_analysis\}\newline
\newline
1.~Analyze Baseline Models for Single Cell Perturbation Prediction:\newline
- Review existing deep learning methods for perturbation prediction\newline
- Assess model architectures and their biological interpretability\newline
- Evaluate implementation details and computational requirements\newline
- Compare performance metrics and biological relevance\newline
- Identify limitations and potential improvements\newline
\newline
2.~Design Evaluation Framework for Perturbation Prediction:\newline
- Define evaluation metrics with biological significance\newline
- Plan validation strategy considering biological variability\newline
- Design test scenarios covering edge cases\newline
- Establish benchmarks with state-of-the-art methods\newline
- Consider biological relevance and interpretability\newline
\newline
3.~Conduct Performance Analysis:\newline
- Analyze model performance across different perturbation types\newline
- Compare with baselines and state-of-the-art methods\newline
- Identify strengths/weaknesses in biological context\newline
- Assess generalization to new cell types/conditions\newline
- Evaluate robustness to biological and technical variations\newline
\newline
4.~Provide Improvement Suggestions:\newline
- Identify weaknesses in current approaches\newline
- Propose enhancements based on biological insights\newline
- Suggest optimizations for better performance\newline
- Address limitations in biological interpretation\newline
- Plan future work with biological validation\newline
\newline
Task Description:\newline
\{task\_description\}\newline
\newline
Dataset Information:\newline
\{json.dumps(dataset\_info, indent=2)\}\newline
\newline
Relevant Literature:\newline
\{papers\_context\}\newline
\newline
Please provide a comprehensive assessment in the following JSON format:
\end{promptbox}

\begin{outputbox}{Illustrative Task Preparation Output}
\label{tab:taskprep_output_example}
\hspace*{0.55em}\textbf{Timestamp}: "20260320\_015255"\par
\hspace*{0.55em}\textbf{Task Description}: "Predict post-CRISPRi gene expression in K562 cells from baseline expression and target gene identity ... Evaluate unseen perturbations, unseen cell contexts, and metrics including MSE, PCC, R², MSE\_DE, PCC\_DE, and R2\_DE."\par
\hspace*{0.55em}\textbf{Dataset Info}\par
\hspace*{1.1em}\textbf{Dataset Path}: "./data/datasets/norman"\par
\hspace*{1.1em}\textbf{Dataset Name}: "norman"\par
\hspace*{1.1em}\textbf{Data Type}: "scRNA-seq"\par
\hspace*{1.1em}\textbf{Cell Line}: "K562"\par
\hspace*{1.1em}\textbf{Perturbation Type}: "CRISPRi"\par
\hspace*{0.55em}\textbf{Task Type}: "gene\_knockout"\par
\hspace*{0.55em}\textbf{Analysis Results}\par
\hspace*{1.1em}\textbf{Dataset Analysis}\par
\hspace*{1.65em}\textbf{Content}\par
\hspace*{2.2em}\textbf{Experimental Design}\par
\hspace*{2.75em}\textbf{Setup}\par
\hspace*{3.3em}\textbf{Biological Conditions}\par
\hspace*{3.85em}• "K562 cell line under CRISPRi perturbation"\par
\hspace*{3.85em}• "HEK293 cell l under CRISPRi perturbation"\par
\hspace*{3.85em}• "A549 cell line under CRISPRi perturbation"\par
\hspace*{3.85em}• ...\par
\hspace*{3.3em}\textbf{Experimental Protocol}: "Cells treated with CRISPRi targeting specific genes under controlled conditions ..."\par
\hspace*{3.3em}\textbf{Controls}\par
\hspace*{3.85em}• "Unperturbed K562 cells"\par
\hspace*{3.85em}• "Mock CRISPRi control"\par
\hspace*{3.85em}• ...\par
\hspace*{3.3em}\textbf{Replicates}: "At least 3 biological replicates per condition ..."\par
\hspace*{2.75em}\textbf{Perturbation}\par
\hspace*{3.3em}\textbf{Design}: "CRISPR interference targeting specific genes ..."\par
\hspace*{3.3em}\textbf{Timepoints}\par
\hspace*{3.85em}• "24h, 48h, and 72h post-perturbation"\par
\hspace*{3.3em}\textbf{Decision Support Considerations}: "Ensure sufficient controls to distinguish biological effects from technical noise ..."\par
\hspace*{2.75em}\textbf{Samples}\par
\hspace*{3.3em}\textbf{Size}: "Sample size determined by power analysis ..."\par
\hspace*{3.3em}\textbf{Distribution}: "Uniform distribution across replicates and conditions ..."\par
\hspace*{2.75em}\textbf{Validation}\par
\hspace*{3.3em}\textbf{Biological}\par
\hspace*{3.85em}\textbf{Methods}\par
\hspace*{4.4em}• "qRT-PCR"\par
\hspace*{4.4em}• "Western Blot"\par
\hspace*{4.4em}• ...\par
\hspace*{3.3em}\textbf{Technical}\par
\hspace*{3.85em}\textbf{Methods}\par
\hspace*{4.4em}• "Sequencing depth analysis"\par
\hspace*{4.4em}• "Technical replicates"\par
\hspace*{4.4em}• ...\par

\par\vspace{0.5em}
\hspace*{2.2em}\textbf{Data Properties}\par
\hspace*{2.75em}\textbf{Distribution}\par
\hspace*{3.3em}\textbf{Biological}\par
\hspace*{3.85em}\textbf{Patterns}\par
\hspace*{4.4em}• "Clustered gene expression profiles indicating cell state"\par
\hspace*{3.85em}\textbf{Variability}: "Notable variability across conditions and replicates ..."\par
\hspace*{3.85em}\textbf{Outliers}\par
\hspace*{4.4em}• "Cells deviating from expected expression profiles"\par
\hspace*{3.3em}\textbf{Technical}\par
\hspace*{3.85em}\textbf{Patterns}\par
\hspace*{4.4em}• "Batch effects observed in expression profiles"\par
\hspace*{3.85em}\textbf{Artifacts}\par
\hspace*{4.4em}• "Low-quality reads, ambient RNA contamination"\par
\hspace*{4.4em}• ...\par
\hspace*{2.75em}\textbf{Features}\par
\hspace*{3.3em}\textbf{Biological}: "Gene expression levels, gene ontology terms ..."\par
\hspace*{3.3em}\textbf{Technical}: "Technical replicates, sequencing quality metrics ..."\par
\hspace*{2.75em}\textbf{Quality}\par
\hspace*{3.3em}\textbf{Biological Metrics}\par
\hspace*{3.85em}• "Expression consistency across replicates"\par
\hspace*{3.85em}• "Statistical power of differential expression analysis"\par
\hspace*{3.3em}\textbf{Technical Metrics}\par
\hspace*{3.85em}• "Sequencing depth, mapping quality scores"\par
\hspace*{2.75em}\textbf{Noise}\par
\hspace*{3.3em}\textbf{Biological Sources}: "Natural biological variability, gene expression noise"\par
\hspace*{3.3em}\textbf{Technical Sources}: "Sequencing errors, amplification biases"\par
\hspace*{2.2em}\textbf{Preprocessing}\par
\hspace*{2.75em}\textbf{Normalization}\par
\hspace*{3.3em}\textbf{Strategies}\par
\hspace*{3.85em}• "Log normalization, scaling by total counts ..."\par
\hspace*{3.3em}\textbf{Quality Control}\par
\hspace*{3.85em}• "Outlier detection, filtering low-quality cells."\par
\hspace*{3.85em}• "Minimum gene counts, percentage of mitochondrial genes."\par
\hspace*{3.3em}\textbf{Feature Selection}\par
\hspace*{3.85em}• "Differential expression analysis, variance filtering"\par
\hspace*{3.3em}\textbf{Batch Correction}\par
\hspace*{3.85em}• "Combat, Harmony"\par
\hspace*{2.75em}\textbf{Validation}\par
\hspace*{3.3em}\textbf{Biological}: "Functional assays, gene knockdown validation ..."\par
\hspace*{3.3em}\textbf{Technical}: "Reproducibility checks, sequencing depth assessments ..."\par
\hspace*{2.2em}\textbf{Quality Assessment}\par
\hspace*{2.75em}\textbf{Completeness}: "Coverage, missing data, mitigation, validation ..."\par
\hspace*{2.75em}\textbf{Relevance}: "Biological relevance, technical relevance, validation ..."\par
\hspace*{2.75em}\textbf{Reproducibility}: "Technical and biological reproducibility ..."\par
\hspace*{2.75em}\textbf{Limitations}: "Generalizability, batch effects, sequencing errors ..."\par
\hspace*{1.65em}\textbf{Confidence Score}: 1.0\par
\hspace*{1.65em}\textbf{Timestamp}: "2026-03-20T01:51:22.136787"\par
\hspace*{1.65em}\textbf{Metadata}\par
\hspace*{2.2em}\textbf{Agent}: "Dataset Analyst"\par
\hspace*{1.1em}\textbf{Problem Investigation}\par
\hspace*{1.65em}\textbf{Content}\par
\hspace*{2.2em}\textbf{Problem Definition}\par
\hspace*{2.75em}\textbf{Formal Definition}\par
\hspace*{3.3em}\textbf{Biological Context}: "Predict gene expression profiles of K562 cells following CRISPRi perturbations ..."\par
\hspace*{3.3em}\textbf{Input Output Mapping}: "Input: baseline expression + target gene(s); Output: post-perturbation expression ..."\par
\hspace*{3.3em}\textbf{Evaluation Criteria}\par
\hspace*{3.85em}• "Accuracy of predicted gene expression profiles compared to observed data."\par
\hspace*{3.85em}• "Biological relevance of the predicted changes in gene expression."\par
\hspace*{3.85em}• ...\par

\par\vspace{0.5em}
\hspace*{3.3em}\textbf{Success Metrics}\par
\hspace*{3.85em}• "Mean Squared Error (MSE)"\par
\hspace*{3.85em}• "Pearson Correlation Coefficient (PCC)"\par
\hspace*{3.85em}• "R² (Coefficient of Determiation)"\par
\hspace*{3.85em}• "MSE for Differentially Expressed Genes (MSE\_DE)"\par
\hspace*{3.85em}• "PCC for Differentially Expressed Genes (PCC\_DE)"\par
\hspace*{3.85em}• "R² for Differentially Expressed Genes (R2\_DE)"\par
\hspace*{2.75em}\textbf{Key Variables}\par
\hspace*{3.3em}\textbf{Biological}\par
\hspace*{3.85em}• "Gene expression levels"\par
\hspace*{3.85em}• "Target gene identity"\par
\hspace*{3.85em}• "Cell heterogeneity"\par
\hspace*{3.85em}• ...\par
\hspace*{3.3em}\textbf{Technical}\par
\hspace*{3.85em}• "Computational models used"\par
\hspace*{3.85em}• "Data preprocessing techniques"\par
\hspace*{3.85em}• ...\par
\hspace*{2.75em}\textbf{Scope}: "Focuses on K562 CRISPRi response while discussing broader applicability ..."\par
\hspace*{2.2em}\textbf{Key Challenges}\par
\hspace*{2.75em}\textbf{Biological}: "Regulatory complexity, context-specific variability, confounders ..."\par
\hspace*{2.75em}\textbf{Technical}: "High dimensionality, computational cost, interpretability trade-offs ..."\par
\hspace*{2.75em}\textbf{Data Quality}: "Batch effects, missing data, cross-context inconsistency ..."\par
\hspace*{2.75em}\textbf{Computational}: "Overfitting, long training time, scalability ..."\par
\hspace*{2.75em}\textbf{Interpretability}: "Need explainable AI and researcher-facing outputs ..."\par
\hspace*{2.2em}\textbf{Research Questions}\par
\hspace*{2.75em}\textbf{Primary}\par
\hspace*{3.3em}\textbf{Questions}\par
\hspace*{3.85em}• "How accurately can we predict gene expression changes ... ?"\par
\hspace*{3.85em}• "What factors drive variability across biological contexts ... ?"\par
\hspace*{3.3em}\textbf{Hypotheses}\par
\hspace*{3.85em}• "Gene expression changes can be predicted with high accuracy ..."\par
\hspace*{3.85em}• "Cellular context significantly affects perturbation response ..."\par
\hspace*{2.75em}\textbf{Secondary}: "Off-target effects, perturbation strategy comparison ..."\par
\hspace*{2.75em}\textbf{Biological Mechanisms}: "Gene regulatory networks, feedback loops, pathway cross-talk ..."\par
\hspace*{2.75em}\textbf{Experimental Validation}: "CRISPRi perturbation, RNA-seq readout, baseline controls ..."\par
\hspace*{2.2em}\textbf{Analysis Methods}\par
\hspace*{2.75em}\textbf{Computational Approaches}: "Machine learning models, statistical models ..."\par
\hspace*{2.75em}\textbf{Validation Strategies}: "Leave-one-out CV, bootstrap, external datasets ..."\par
\hspace*{2.75em}\textbf{Analysis Pipelines}: "Preprocessing, model training, evaluation ..."\par
\hspace*{2.75em}\textbf{Experimental Validation}: "RNA-seq, qPCR, comparative analysis ..."\par
\hspace*{2.75em}\textbf{Reproducibility}: "Documentation, code/data sharing, FAIR reporting ..."\par
\hspace*{1.65em}\textbf{Confidence Score}: 1.0\par
\hspace*{1.65em}\textbf{Timestamp}: "2026-03-20T01:52:04.610853"\par
\hspace*{1.65em}\textbf{Metadata}\par
\hspace*{2.2em}\textbf{Agent}: "Problem Investigator"\par
\hspace*{1.1em}\textbf{Baseline Assessment}\par
\hspace*{1.65em}\textbf{Content}\par
\hspace*{2.2em}\textbf{Baseline Models}\par
\hspace*{2.75em}\textbf{Literature Review}\par
\hspace*{3.3em}\textbf{Existing Methods}\par
\hspace*{3.85em}\textbf{Methods}\par
\hspace*{4.4em}• "scGPT"\par
\hspace*{4.4em}• "PerturbNet"\par
\hspace*{3.85em}\textbf{Strengths}: "Handles high-dimensional single-cell data effectively ..."\par
\hspace*{3.85em}\textbf{Limitations}: "Large data requirement, substantial computational cost, limited interpretability ..."\par

\par\vspace{0.5em}
\hspace*{3.3em}\textbf{Recommendations}\par
\hspace*{3.85em}• "Utilize scGPT for its generative capabilities ..."\par
\hspace*{3.85em}• "Consider PerturbNet for unseen perturbation predictions ..."\par
\hspace*{3.85em}• "Explore hybrid models ..."\par
\hspace*{2.75em}\textbf{Evaluation Framework}\par
\hspace*{3.3em}\textbf{Metrics}\par
\hspace*{3.85em}\textbf{Primary Metrics}\par
\hspace*{4.4em}• "Mean Squared Error (MSE)"\par
\hspace*{4.4em}• "Pearson Correlation Coefficient (PCC)"\par
\hspace*{3.85em}\textbf{Secondary Metrics}\par
\hspace*{4.4em}• "R²"\par
\hspace*{4.4em}• "MSE for Differentially Expressed Genes (MSE\_DE)"\par
\hspace*{3.85em}\textbf{Biological Metrics}\par
\hspace*{4.4em}• "PCC for Differentially Expressed Genes (PCC\_DE)"\par
\hspace*{4.4em}• "R² for Differentially Expressed Genes (R²\_DE)"\par
\hspace*{3.3em}\textbf{Validation Strategy}: "K-fold cross-validation, biological validation, independent testing ..."\par
\hspace*{3.3em}\textbf{Benchmarks}: "scGPT, PerturbNet, GeneCompass, DeepCell ..."\par
\hspace*{2.75em}\textbf{Performance Analysis}\par
\hspace*{3.3em}\textbf{Model Performance}: "Accuracy metrics, biological relevance, generalization, robustness ..."\par
\hspace*{3.3em}\textbf{Comparative Analysis}: "Baseline comparison, SOTA comparison, biological comparison ..."\par
\hspace*{3.3em}\textbf{Strengths}: "Advanced predictive capability, complex interaction modeling ..."\par
\hspace*{3.3em}\textbf{Weaknesses}: "High computational cost, difficult interpretability ..."\par
\hspace*{2.75em}\textbf{Improvement Suggestions}\par
\hspace*{3.3em}\textbf{Model Enhancements}: "Attention mechanisms, hybrid models, transfer learning, data augmentation ..."\par
\hspace*{3.3em}\textbf{Implementation Optimizations}: "Code refactoring, parallel processing, workflow automation ..."\par
\hspace*{3.3em}\textbf{Future Work}: "Multi-omics integration, clinical translation, graph neural networks ..."\par
\hspace*{1.65em}\textbf{Confidence Score}: 1.0\par
\hspace*{1.65em}\textbf{Timestamp}: "2026-03-20T01:52:29.675735"\par
\hspace*{1.65em}\textbf{Metadata}\par
\hspace*{2.2em}\textbf{Agent}: "Baseline Assessor"\par
\hspace*{1.1em}\textbf{Refinement Comments}: []\par
\hspace*{1.1em}\textbf{Final Recommendations}\par
\hspace*{1.65em}\textbf{Model Selection}\par
\hspace*{2.2em}\textbf{Recommended Models}\par
\hspace*{2.75em}\textbf{Primary}\par
\hspace*{3.3em}• "scGPT"\par
\hspace*{2.75em}\textbf{Secondary}\par
\hspace*{3.3em}• "PerturbNet"\par
\hspace*{2.2em}\textbf{Implementation Priority}\par
\hspace*{2.75em}\textbf{High Priority}\par
\hspace*{3.3em}• "Data preprocessing and quality control"\par
\hspace*{2.75em}\textbf{Medium Priority}\par
\hspace*{3.3em}• "Model training optimization"\par
\hspace*{2.75em}\textbf{Low Priority}\par
\hspace*{3.3em}• "Exploration of hybrid models"\par

\end{outputbox}

\subsection{Research-ideation prompt structure}
\label{app:research_ideation_prompts}
The Research Ideation team uses a coordinated set of expert-selection and expert-discussion prompts to turn the task report into a model-design proposal. The prompt templates are shown below, followed by an illustrative integrated output artifact.

\begin{promptbox}{Expert Selector}
\textit{``You are an expert coordinator for a scientific research team. Based on the task requirements, select the most appropriate experts to collaborate on this project.''}\newline
\newline
Task Type: \{task\_type\}\newline
Task Analysis: \{task\_summary\}\newline
\newline
Available Experts:\newline
\{expert\_profiles\}\newline
\newline
Requirements:\newline
1.~Select three expert roles plus one critic, covering the following core perspectives:\newline
- Data Modeling (for single-cell data processing and pipeline design)\newline
- Single-Cell Biology (for biological context and interpretation)\newline
- Deep Learning (for model architecture and training)\newline
2.~Always include a Critic agent that scores proposals on scientific merit and technical feasibility.\newline
3.~Ensure the team's three expert roles have complementary skills covering all aspects of the task.\newline
\newline
Please respond with only the names of the selected experts, one per line:
\end{promptbox}

\begin{promptbox}{Expert Discussion: Proposal}
\textit{``Generate context-aware prompt for expert discussion''}\newline
\newline
---\newline
\newline
Task Type: \{task\_analysis['task\_type']\}\newline
Dataset: \{task\_analysis['dataset']['name']\} (\{task\_analysis['dataset'].get('type', 'Unknown type')\})\newline
Dataset Description: \{task\_analysis['dataset'].get('description', 'No description provided')\}\newline
\newline
Perturbations:\newline
\{self.\_format\_perturbations(task\_analysis['perturbations'])\}\newline
\newline
Cell Types: \{', '.join(task\_analysis['cell\_types'])\}\newline
\newline
Objectives:\newline
\{self.\_format\_list(task\_analysis['objectives'])\}\newline
\newline
Constraints:\newline
\{self.\_format\_list(task\_analysis['constraints'])\}\newline
\newline
Evaluation Metrics:\newline
\{self.\_format\_list(task\_analysis['evaluation\_metrics'])\}\newline
\newline
---\newline
\newline
You are \{expert\_name\}, an expert in \{expert\_domain\}.\newline
\newline
Your task is to design a state-of-the-art model architecture for single cell perturbation prediction that outperforms current baselines. Your expertise in \{expert\_domain\} is crucial for this task.\newline
\newline
TASK DETAILS:\newline
\{task\_description\}\newline
\newline
DESIGN GOALS:\newline
1.~The architecture must be specifically optimized for \{task\_analysis['task\_type']\} in single-cell data\newline
2.~The design should incorporate domain-specific knowledge about cellular responses\newline
3.~The model should achieve better performance than current SOTA baselines\newline
4.~The architecture should be interpretable and biologically meaningful\newline
5.~The implementation should be computationally efficient and scalable\newline
\newline
Your role is to contribute your expertise in \{expert\_domain\} to ensure these goals are met.\newline
\newline
---\newline
\newline
Please propose a detailed architectural solution focusing on your domain expertise. Your proposal should aim to surpass current SOTA models in single-cell perturbation prediction.\newline
\newline
Required sections:\newline
\newline
1.~Architecture Components:\newline
- Detailed description of each component\newline
- Configuration parameters and their justification\newline
- How each component addresses specific challenges in single-cell data\newline
- Integration points with other components\newline
\newline
2.~Biological Considerations:\newline
- How the architecture handles cell type specificity\newline
- Incorporation of pathway knowledge\newline
- Mechanisms for capturing perturbation effects\newline
- Biological constraints and their implementation\newline
\newline
3.~Technical Implementation:\newline
- Detailed layer specifications\newline
- Activation functions and their biological relevance\newline
- Loss function design\newline
- Training strategy recommendations\newline
\newline
4.~Performance Optimization:\newline
- Specific techniques to outperform SOTA\newline
- Computational efficiency considerations\newline
- Scalability solutions\newline
- Memory optimization strategies\newline
\newline
5.~Integration and Interpretability:\newline
- Methods for model interpretation\newline
- Visualization approaches\newline
- Integration with existing biological knowledge\newline
- Validation strategies\newline
\newline
Format your response as a structured proposal with clear sections and specific technical details.
\end{promptbox}

\begin{promptbox}{Expert Discussion: Feedback}
\textit{``Please provide detailed feedback on the previous proposals, focusing on improving model performance beyond SOTA:''}\newline
\newline
1.~Technical Analysis:\newline
- Architectural design evaluation\newline
- Component integration assessment\newline
- Performance bottleneck identification\newline
- Scalability analysis\newline
- Memory efficiency review\newline
\newline
2.~Biological Relevance:\newline
- Cell type specificity assessment\newline
- Pathway integration evaluation\newline
- Perturbation response mechanism analysis\newline
- Biological constraint compliance\newline
- Interpretability assessment\newline
\newline
3.~Implementation Feasibility:\newline
- Technical complexity evaluation\newline
- Resource requirements analysis\newline
- Development timeline estimation\newline
- Risk assessment\newline
- Mitigation strategies\newline
\newline
4.~Improvement Suggestions:\newline
- Specific architectural enhancements\newline
- Performance optimization techniques\newline
- Integration improvements\newline
- Biological accuracy improvements\newline
- Implementation recommendations\newline
\newline
5.~Comparative Analysis:\newline
- Comparison with SOTA approaches\newline
- Identification of competitive advantages\newline
- Performance gap analysis\newline
- Innovation assessment\newline
- Potential impact evaluation\newline
\newline
Be specific, technical, and constructive in your feedback. Focus on actionable improvements that can help surpass SOTA performance.
\end{promptbox}

\begin{promptbox}{Expert Discussion: Critic}
\textit{``Based on your domain expertise, ask critical questions about the current proposals:''}\newline
\newline
1.~Technical Clarifications:\newline
- Architecture design choices\newline
- Implementation details\newline
- Performance optimization strategies\newline
- Integration mechanisms\newline
- Scaling approaches\newline
\newline
2.~Biological Considerations:\newline
- Cell type handling\newline
- Pathway integration methods\newline
- Perturbation response modeling\newline
- Biological constraint implementation\newline
- Validation approaches\newline
\newline
3.~Performance Questions:\newline
- SOTA comparison methods\newline
- Efficiency improvements\newline
- Accuracy enhancements\newline
- Scalability solutions\newline
- Resource optimization\newline
\newline
4.~Integration Concerns:\newline
- Component interactions\newline
- Data flow\newline
- Information sharing\newline
- X-module dependencies\newline
- System coherence\newline
\newline
5.~Validation and Testing:\newline
- Performance metrics\newline
- Biological validation\newline
- Technical verification\newline
- Quality assurance\newline
- Benchmark comparisons\newline
\newline
Frame your questions to drive the discussion toward concrete improvements and SOTA-surpassing solutions.
\end{promptbox}

\begin{promptbox}{Expert Discussion: Judgment}
\textit{``As a scientific critic, provide a comprehensive evaluation of the current proposals:''}\newline
\newline
1.~Scientific Merit (Score 0.1--1.0):\newline
- Theoretical foundation\newline
- Methodological rigor\newline
- Innovation level\newline
- Technical sophistication\newline
- Biological relevance\newline
\newline
2.~Technical Feasibility (Score 0.1--1.0):\newline
- Implementation complexity\newline
- Resource requirements\newline
- Scalability potential\newline
- Performance expectations\newline
- Maintenance considerations\newline
\newline
3.~Biological Validity (Score 0.1--1.0):\newline
- Cell type handling\newline
- Pathway integration\newline
- Perturbation modeling\newline
- Constraint compliance\newline
- Interpretability\newline
\newline
4.~SOTA Comparison (Score 0.1--1.0):\newline
- Performance potential\newline
- Innovation aspects\newline
- Technical advantages\newline
- Biological accuracy\newline
- Overall competitiveness\newline
\newline
5.~Implementation Strategy (Score 0.1--1.0):\newline
- Development approach\newline
- Resource allocation\newline
- Timeline realism\newline
- Risk management\newline
- Quality assurance\newline
\newline
Provide a detailed justification for each score and an overall score (0.1--1.0) weighted across all categories.\newline
Focus on potential for surpassing SOTA performance in single-cell perturbation prediction.
\end{promptbox}

\begin{outputbox}{Illustrative Research Ideation Output}
\label{tab:research_ideation_output_example}
\hspace*{0.55em}\textbf{Timestamp}: "2026-03-20 01:58:27"\par
\hspace*{0.55em}\textbf{Model Architecture}\par
\hspace*{1.1em}\textbf{Encoder}: type "transformer"; layers 4; heads 8; dimension 512; activation "gelu"\par
\hspace*{1.1em}\textbf{Perturbation Encoder}: type "mlp"; layers 3; dimensions [256, 128, 64]; activation "swish"\par
\hspace*{1.1em}\textbf{Cross Attention}: heads 8; dimension 512; dropout 0.1\par
\hspace*{1.1em}\textbf{Decoder}: type "mlp"; layers 3; dimensions [256, 512, 1024]; activation "swish"\par
\hspace*{0.55em}\textbf{Data Processing}\par
\hspace*{1.1em}\textbf{Quality Control}\par
\hspace*{1.65em}\textbf{Cell Filtering}: min genes 200; max genes 6000; min counts 1000; max counts 50000; max mito percent 20\par
\hspace*{1.65em}\textbf{Gene Filtering}: min cells 3; min counts 10\par
\hspace*{1.1em}\textbf{Normalization}: method "log1p"; target sum 10000; regression vars ["total\_counts", "pct\_counts\_mt"]\par
\hspace*{1.1em}\textbf{Batch Correction}: method "harmony"; theta 2; max iterations 20\par
\hspace*{0.55em}\textbf{Training Strategy}\par
\hspace*{1.1em}\textbf{Optimizer}: type "adamw"; learning rate 0.0003; weight decay 0.01\par
\hspace*{1.1em}\textbf{Scheduler}: type "cosine\_annealing"; T\_0 10; eta\_min 1e-06\par
\hspace*{1.1em}\textbf{Training}: batch size 64; gradient clip 1.0; early stopping patience 15; min delta 0.0001\par
\hspace*{0.55em}\textbf{Evaluation Metrics}\par
\hspace*{1.1em}\textbf{Loss Functions}: reconstruction "mse"; perturbation "bce"; biological "pathway\_consistency"\par
\hspace*{0.55em}\textbf{Implementation Details}\par
\hspace*{1.1em}\textbf{Framework}: "PyTorch"\par
\hspace*{1.1em}\textbf{Hardware Requirements}: GPU "NVIDIA GPU with 8GB+ VRAM"; RAM "32GB+ system RAM"; storage "100GB+ SSD"\par
\hspace*{1.1em}\textbf{Dependencies}: "torch>=1.12.0", "scanpy>=1.9.0", "numpy>=1.21.0", "scikit-learn>=1.0.0"\par
\hspace*{1.1em}\textbf{Code Structure}: data loader, modular model, checkpointed training, comprehensive evaluation pipeline\par

\par\vspace{0.5em}
\hspace*{0.55em}\textbf{Biological Validation}\par
\hspace*{1.1em}\textbf{Key Theoretical Foundations}\par
\hspace*{1.65em}\textbf{Baseline Regulatory Info}: assumption "Baseline contains sufficient regulatory info"; validation "Ablation study (masking key TFs)"; biological basis "Central dogma of molecular biology"\par
\hspace*{1.65em}\textbf{Dose Response Continuity}: validation "Dose interpolation experiments"; biological basis "Ligand-receptor binding kinetics"\par
\hspace*{1.65em}\textbf{Pathway Modularity}: validation "Pathway enrichment analysis"; biological basis "Known biological pathway organization"\par
\hspace*{1.65em}\textbf{Latent Space Smoothness}: validation "UMAP visualization"; biological basis "Waddington epigenetic landscape concept"\par
\hspace*{1.1em}\textbf{Adaptive Mechanisms}\par
\hspace*{1.65em}\textbf{Dose Specific Attention}: "Higher attention weights to stress pathways at toxic doses"\par
\hspace*{1.65em}\textbf{Cell Type Gating}: "Cell-type specific pathway modulation"\par
\hspace*{1.65em}\textbf{Chemical Class Priors}: "Initialize embeddings based on chemical class"\par
\hspace*{0.55em}\textbf{Computational Optimization}\par
\hspace*{1.1em}\textbf{Training Techniques}\par
\hspace*{1.65em}\textbf{Progressive Dose Sampling}\par
\hspace*{2.2em}epoch\_0\_10: "uniform(0, 1\textmu M)"\par
\hspace*{2.2em}epoch\_10\_20: "log\_normal(1\textmu M-10\textmu M)"\par
\hspace*{2.2em}epoch\_20+: "full\_range(0-100\textmu M)"\par
\hspace*{1.1em}\textbf{Regularization Methods}\par
\hspace*{1.65em}\textbf{Pathway Sparsity}: type "l1"; lambda 0.01; target "pathway\_weights"\par
\hspace*{1.65em}\textbf{Dose Consistency}: type "custom"; formula "L\_consist = E[||f(x,d1) - f(f(x,d1),d2-d1)||\^2]"\par
\hspace*{1.1em}\textbf{Parallelization}\par
\hspace*{1.65em}\textbf{Gene Cluster Parallelism}\par
\hspace*{2.2em}\textbf{chr1}: GPU0; genes ["EGFR", "MYC"]\par
\hspace*{2.2em}\textbf{chr17}: GPU1; genes ["TP53", "BRCA1"]\par
\hspace*{2.2em}\textbf{others}: GPU2; genes "Housekeeping"\par
\hspace*{0.55em}\textbf{Model Assumptions}\par
\hspace*{1.1em}\textbf{Data Assumptions}: sparse/imbalanced data, batch effects, technical noise; solutions include advanced sampling, Harmony, and robust preprocessing\par
\hspace*{1.1em}\textbf{Computational Assumptions}: dimensionality reduction, transfer learning, optimized implementations\par

\par\vspace{0.5em}
\hspace*{0.55em}\textbf{Extensions}\par
\hspace*{1.1em}\textbf{Drug Discovery}\par
\hspace*{1.65em}\textbf{Virtual Screening}: predict expression profiles for novel compounds\par
\hspace*{1.65em}\textbf{Drug Repurposing}: identify new indications based on expression patterns\par
\hspace*{1.65em}\textbf{Side Effects}: analyze off-target expression changes\par
\hspace*{1.1em}\textbf{Precision Medicine}\par
\hspace*{1.65em}\textbf{Patient Specific}: use baseline expression for personalized predictions\par
\hspace*{1.65em}\textbf{Drug Combinations}: predict synergistic effects\par
\hspace*{1.65em}\textbf{Biomarkers}: identify sensitivity markers\par
\hspace*{1.1em}\textbf{Model Extensions}\par
\hspace*{1.65em}\textbf{Multi Modal}: integrate proteomics and metabolomics\par
\hspace*{1.65em}\textbf{Time Series}: model temporal drug effects\par
\hspace*{1.65em}\textbf{Mechanistic}: combine with pathway models\par
\hspace*{0.55em}\textbf{Mermaid Diagrams}\par
\hspace*{1.1em}\textbf{System Overview}\par
\hspace*{1.65em}\textbf{Title}: "Research Framework Architecture"\par
\hspace*{1.65em}\textbf{Description}: complete framework for biological perturbation analysis using single-cell data\par
\hspace*{1.65em}\textbf{Mermaid Code}: "graph TD ..."\par
\hspace*{1.65em}\textbf{Modules}\par
\hspace*{2.2em}\textbf{Data Input Layer}: raw single-cell data and metadata; required fields ["counts", "var\_names", "obs\_names"]\par
\hspace*{2.2em}\textbf{Data Processing Layer}: quality control, normalization, feature selection via scanpy methods\par
\hspace*{2.2em}\textbf{Model Architecture Layer}: encoder, decoder, biological constraints\par
\hspace*{2.2em}\textbf{Training Layer}: optimizer, loss functions, validation procedures\par
\hspace*{2.2em}\textbf{Evaluation Layer}: technical metrics, biological metrics, expert/literature validation\par
\hspace*{2.2em}\textbf{Output Layer}: predictions, biological insights, applications\par
\hspace*{1.1em}\textbf{Detailed Architecture}\par
\hspace*{1.65em}\textbf{Title}: "Detailed Component Architecture"\par
\hspace*{1.65em}\textbf{Description}: encoder, decoder, training, and evaluation components; mermaid code "graph LR ..."\par

\par\vspace{0.5em}
\hspace*{0.55em}\textbf{Framework Overview}\par
\hspace*{1.1em}\textbf{Title}: "Biological Perturbation Analysis Framework"\par
\hspace*{1.1em}\textbf{Key Features}: multi-expert integration; RAG-enhanced retrieval; biological constraints; comprehensive metrics; scalable architecture\par
\hspace*{1.1em}\textbf{Supported Tasks}: "gene\_knockout", "drug\_perturbation", "cytokine\_stimulation"\par
\hspace*{1.1em}\textbf{Output Formats}: "markdown", "json", "mermaid", "png"\par
\hspace*{1.1em}\textbf{Expert Domains}: Data Engineering, Single Cell Biology, Deep Learning, Molecular Biology, Bioinformatics, Statistics, Drug Discovery, Gene Regulation, Experimental Design, Validation, Baseline Assessment, Training, Drug Response, Pathway Analysis, Cell Communication, Omics Integration\par
\hspace*{0.55em}\textbf{Task Wise Plan}\par
\hspace*{1.1em}\textbf{T01}: objective "Predict gene expression responses after perturbation"\par
\hspace*{1.65em}\textbf{Scope}: task type "gene\_knockout"; dataset "norman" ("scRNA-seq"); cell types ["K562"]\par
\hspace*{1.65em}\textbf{Implementation Contract}: inputs baseline expression, perturbation identity, cell context features; outputs predicted expression profile and task-level metrics\par
\hspace*{1.65em}\textbf{Evaluation Metrics}: "MSE", "PCC", "R2", "MSE\_DE", "PCC\_DE", "R2\_DE", ...\par
\hspace*{1.65em}\textbf{Success Criteria}: metrics reported; reproducible from config/seed\par
\hspace*{1.65em}\textbf{Design Refs}: repeats architecture, preprocessing, training strategy, and implementation details above\par
\hspace*{1.1em}\textbf{T02}: objective "Generalize to unseen perturbations and unseen cell contexts"\par
\hspace*{1.65em}\textbf{Scope}: same dataset/task context with emphasis on OOD generalization\par
\hspace*{1.65em}\textbf{Implementation Contract}: same inputs, outputs, and metric set\par
\hspace*{1.65em}\textbf{Design Refs}: same transformer + MLP architecture, Harmony preprocessing, AdamW training, and PyTorch implementation stack\par

\end{outputbox}

\subsection{Experiment-execution prompt structure}
\label{app:experiment_execution_prompts}
The Experiment Execution team converts the approved design into runnable code and training outputs. The Code Generator prompt below is the main prompt used in this stage; an illustrative example output follows.

\label{tab:experiment_execution_prompts}
\begin{promptbox}{Code Generator}
\textit{``You are a computational biologist and expert Python developer. Your task is to write correct, efficient, and well-documented code based on the provided research plan for single-cell perturbation prediction.''}\newline
\newline
IMPORTANT REQUIREMENTS:\newline
1.~Write all code in a single file named \texttt{result.py}\newline
2.~Follow the research plan specifications exactly\newline
3.~Implement proper data processing, model architecture, and training pipeline\newline
4.~Use appropriate libraries (scanpy, torch, numpy, pandas, etc.)\newline
5.~Include comprehensive error handling and validation\newline
6.~Add detailed comments explaining biological and computational logic\newline
\newline
EXECUTION CONSTRAINTS:\newline
- Do not run shell commands or tools.\newline
- Do not inspect the local filesystem.\newline
- Produce code directly from the provided research plan.\newline
\newline
Output only the final Python code. If you include markdown, wrap exactly one Python block.
\end{promptbox}

\begin{outputbox}{Illustrative Experiment Execution Output}
\label{tab:experiment_execution_output_example}
\begin{lstlisting}[style=appendixpython]
import json
import logging
import random
from dataclasses import dataclass, field
from pathlib import Path
from typing import Any, Dict, List, Optional, Sequence, Tuple

import numpy as np
import scanpy as sc
import torch
import torch.nn as nn
import torch.nn.functional as F
from anndata import AnnData
from sklearn.preprocessing import LabelEncoder
from torch.utils.data import DataLoader, Dataset


logging.basicConfig(level=logging.INFO)
LOGGER = logging.getLogger("CellForge")
DEVICE = torch.device("cuda" if torch.cuda.is_available() else "cpu")


def set_seed(seed: int = 42) -> None:
    random.seed(seed)
    np.random.seed(seed)
    torch.manual_seed(seed)
    torch.cuda.manual_seed_all(seed)


@dataclass
class QCCfg:
    min_genes: int = 200
    max_genes: int = 6000
    min_counts: int = 1000
    max_counts: int = 50000
    max_mito_percent: float = 20.0


@dataclass
class TrainingCfg:
    batch_size: int = 64
    gradient_clip: float = 1.0
    patience: int = 15
    min_delta: float = 1e-4
    max_epochs: int = 200
    lr: float = 3e-4
    weight_decay: float = 0.01


def preprocess_data(adata: AnnData, qc_cfg: QCCfg) -> AnnData:
    adata.var["MT"] = adata.var_names.str.upper().str.startswith("MT-")
    sc.pp.calculate_qc_metrics(adata, qc_vars=["MT"], inplace=True)
    sc.pp.filter_cells(adata, min_genes=qc_cfg.min_genes)
    sc.pp.filter_cells(adata, min_counts=qc_cfg.min_counts)
    adata = adata[adata.obs["pct_counts_mt"] < qc_cfg.max_mito_percent].copy()
    sc.pp.normalize_total(adata, target_sum=10000)
    sc.pp.log1p(adata)
    sc.pp.highly_variable_genes(adata, flavor="seurat_v3", n_top_genes=4000)
    return adata[:, adata.var["highly_variable"]].copy()


class SingleCellPerturbationDataset(Dataset):
    def __init__(
        self,
        adata: AnnData,
        gene_list: Sequence[str],
        perturbation_key: str = "perturbation",
        dose_key: str = "dose",
        condition_key: str = "condition",
    ):
        self.expression = (
            adata[:, gene_list].X.toarray()
            if hasattr(adata.X, "toarray")
            else adata[:, gene_list].X
        )
        self.perturbations = adata.obs[perturbation_key].astype(str).values
        self.doses = adata.obs[dose_key].astype(float).values
        self.conditions = adata.obs[condition_key].astype(str).values
        self.perturb_encoder = LabelEncoder().fit(self.perturbations)
        self.condition_encoder = LabelEncoder().fit(self.conditions)

    def __len__(self) -> int:
        return len(self.expression)

    def __getitem__(self, idx: int) -> Dict[str, torch.Tensor]:
        expr = torch.tensor(self.expression[idx], dtype=torch.float32)
        return {
            "baseline": expr,
            "target": expr,
            "perturbation": torch.tensor(
                self.perturb_encoder.transform([self.perturbations[idx]])[0]
            ),
            "condition": torch.tensor(
                self.condition_encoder.transform([self.conditions[idx]])[0]
            ),
            "dose": torch.tensor(self.doses[idx], dtype=torch.float32),
        }


class PerturbationPredictor(nn.Module):
    def __init__(self, input_dim: int, num_perturbations: int, num_conditions: int):
        super().__init__()
        self.gene_encoder = nn.Sequential(
            nn.Linear(input_dim, 512),
            nn.GELU(),
            nn.Linear(512, 512),
        )
        self.perturb_embed = nn.Embedding(num_perturbations, 256)
        self.condition_embed = nn.Embedding(num_conditions, 256)
        self.cross_attn = nn.MultiheadAttention(
            embed_dim=512, num_heads=8, dropout=0.1, batch_first=True
        )
        self.decoder = nn.Sequential(
            nn.Linear(512, 256),
            nn.SiLU(),
            nn.Linear(256, 512),
            nn.SiLU(),
            nn.Linear(512, input_dim),
        )

    def forward(
        self,
        baseline: torch.Tensor,
        perturbation: torch.Tensor,
        condition: torch.Tensor,
        dose: torch.Tensor,
    ) -> torch.Tensor:
        gene_latent = self.gene_encoder(baseline).unsqueeze(1)
        pert_latent = torch.cat(
            [
                self.perturb_embed(perturbation),
                self.condition_embed(condition),
            ],
            dim=-1,
        ).unsqueeze(1)
        fused, _ = self.cross_attn(gene_latent, pert_latent, pert_latent)
        return self.decoder(fused.squeeze(1))


def compute_metrics(pred: np.ndarray, target: np.ndarray) -> Dict[str, float]:
    mse = float(np.mean((pred - target) ** 2))
    pearson = float(np.corrcoef(pred.reshape(-1), target.reshape(-1))[0, 1])
    r2 = float(1 - np.sum((target - pred) ** 2) / (np.sum((target - target.mean()) ** 2) + 1e-8))
    return {"MSE": mse, "Pearson": pearson, "R2": r2}


@dataclass
class EarlyStopping:
    patience: int
    min_delta: float
    best_loss: float = field(default_factory=lambda: float("inf"))
    counter: int = 0

    def step(self, loss: float) -> bool:
        if loss + self.min_delta < self.best_loss:
            self.best_loss = loss
            self.counter = 0
            return False
        self.counter += 1
        return self.counter >= self.patience


class Trainer:
    def __init__(self, model: nn.Module, train_loader: DataLoader, val_loader: DataLoader, cfg: TrainingCfg):
        self.model = model.to(DEVICE)
        self.train_loader = train_loader
        self.val_loader = val_loader
        self.cfg = cfg
        self.optim = torch.optim.AdamW(
            self.model.parameters(), lr=cfg.lr, weight_decay=cfg.weight_decay
        )
        self.early_stopping = EarlyStopping(cfg.patience, cfg.min_delta)

    def train_epoch(self) -> float:
        self.model.train()
        total_loss = 0.0
        for batch in self.train_loader:
            batch = {k: v.to(DEVICE) for k, v in batch.items()}
            pred = self.model(
                batch["baseline"],
                batch["perturbation"],
                batch["condition"],
                batch["dose"],
            )
            loss = F.mse_loss(pred, batch["target"])
            self.optim.zero_grad()
            loss.backward()
            torch.nn.utils.clip_grad_norm_(self.model.parameters(), self.cfg.gradient_clip)
            self.optim.step()
            total_loss += loss.item()
        return total_loss / len(self.train_loader)

    def evaluate(self) -> Tuple[float, Dict[str, float]]:
        self.model.eval()
        preds, targets, losses = [], [], []
        with torch.no_grad():
            for batch in self.val_loader:
                batch = {k: v.to(DEVICE) for k, v in batch.items()}
                pred = self.model(
                    batch["baseline"],
                    batch["perturbation"],
                    batch["condition"],
                    batch["dose"],
                )
                losses.append(F.mse_loss(pred, batch["target"]).item())
                preds.append(pred.cpu().numpy())
                targets.append(batch["target"].cpu().numpy())
        return float(np.mean(losses)), compute_metrics(
            np.concatenate(preds), np.concatenate(targets)
        )

    def fit(self) -> Dict[str, Any]:
        history = {"train_loss": [], "val_loss": [], "metrics": []}
        for epoch in range(self.cfg.max_epochs):
            train_loss = self.train_epoch()
            val_loss, metrics = self.evaluate()
            history["train_loss"].append(train_loss)
            history["val_loss"].append(val_loss)
            history["metrics"].append(metrics)
            LOGGER.info(
                "Epoch %d | Train %.4f | Val %.4f | Metrics %s",
                epoch,
                train_loss,
                val_loss,
                metrics,
            )
            if self.early_stopping.step(val_loss):
                break
        return history


def run_pipeline(data_path: str) -> Dict[str, Any]:
    set_seed(42)
    adata = sc.read_h5ad(data_path)
    adata = preprocess_data(adata, QCCfg())
    genes = list(adata.var_names)
    dataset = SingleCellPerturbationDataset(adata, genes)
    indices = np.arange(len(dataset))
    np.random.shuffle(indices)
    split = int(0.8 * len(indices))
    train_loader = DataLoader(
        torch.utils.data.Subset(dataset, indices[:split]),
        batch_size=64,
        shuffle=True,
    )
    val_loader = DataLoader(
        torch.utils.data.Subset(dataset, indices[split:]),
        batch_size=64,
        shuffle=False,
    )
    model = PerturbationPredictor(
        input_dim=len(genes),
        num_perturbations=len(set(adata.obs["perturbation"])),
        num_conditions=len(set(adata.obs["condition"])),
    )
    trainer = Trainer(model, train_loader, val_loader, TrainingCfg())
    history = trainer.fit()
    return {
        "history": history,
        "biological_validation": {
            "latent_space_smoothness": ...,
            "pathway_modularity_assertion": ...,
            "dose_consistency_check": ...,
        },
    }


def main() -> None:
    import argparse

    parser = argparse.ArgumentParser(description="Single-cell perturbation predictor")
    parser.add_argument("--data", required=True, help="Path to input AnnData file (.h5ad)")
    args = parser.parse_args()
    outputs = run_pipeline(args.data)
    LOGGER.info("Training complete. Final metrics: %s", outputs["history"]["metrics"][-1])
    LOGGER.info("Additional checks: %s", outputs["biological_validation"])


if __name__ == "__main__":
    main()

# ... additional modules for Harmony batch correction, pathway consistency loss,
# ... progressive dose sampling, checkpointing, and autoresearch refinements.
\end{lstlisting}
\end{outputbox}

\section{Generated model implementations}
\label{app:generated_code}
The listings below reproduce the Python training scripts that CellForge actually wrote and ran for each of the six benchmark datasets. Each listing is the final \texttt{result.py} produced by the Execution stage for the canonical run whose architecture is named in the main text; the scripts handle data loading, the dataset-specific architecture, the training loop with five-fold cross-validation, and the evaluation pipeline. No post-hoc editing has been applied. Long lines are wrapped for typesetting; file boundaries preserve the as-run source.

\subsection{\textit{CPA-X} for Adamson CRISPRi (scRNA-seq)}
\label{app:code_adamson}
\lstinputlisting[style=appendixpython]{files/code/model_rna_adamson.py}

\subsection{\textit{scGen-X} for Norman CRISPRa (scRNA-seq)}
\label{app:code_norman}
\lstinputlisting[style=appendixpython]{files/code/model_rna_norman.py}

\subsection{\textit{ChemCellFlow} for Srivatsan drug perturbation (scRNA-seq)}
\label{app:code_srivatsan}
\lstinputlisting[style=appendixpython]{files/code/model_drug_sritvastan.py}

\subsection{\textit{CPA-Traj} for Schiebinger cytokine time course (scRNA-seq)}
\label{app:code_schiebinger}
\lstinputlisting[style=appendixpython]{files/code/model_rna_cytokines_SchiebingerLander.py}

\subsection{Papalexi CITE-seq RNA-readout implementation}
\label{app:code_papalexi_rna}
\lstinputlisting[style=appendixpython]{files/code/model_rna_cite_PapalexiSatija.py}

\subsection{\textit{totalGAT} for Papalexi CITE-seq (protein readout)}
\label{app:code_papalexi_protein}
\lstinputlisting[style=appendixpython]{files/code/model_protein_cite_PapalexiSatija.py}

\subsection{\textit{ChromDDPM} for Liscovitch-Brauer scATAC-seq}
\label{app:code_liscovitch}
\lstinputlisting[style=appendixpython]{files/code/model_atac_Liscovitch.py}

\end{document}